\documentclass{article}

 \usepackage{iclr_2026}

\usepackage[utf8]{inputenc} 
\usepackage[T1]{fontenc}    
\usepackage{hyperref}       
\usepackage{url}            
\usepackage{booktabs}       
\usepackage{amsfonts}       
\usepackage{nicefrac}       
\usepackage{microtype}      
\usepackage{xcolor}         

\usepackage[page,title]{appendix}
\usepackage{comment}
\usepackage{enumitem}
\usepackage{threeparttable}
\usepackage{tikz}
\usetikzlibrary{positioning, arrows.meta, calc}

\usepackage{multirow}
\usepackage{float}

\usepackage{bm}
\usepackage{dsfont}
\usepackage{amsmath}
\usepackage{amssymb}
\usepackage{amsthm}
    
\newtheorem{theorem}{Theorem}  

\newtheorem{condition}{Condition}
\newtheorem{assumption}{Assumption}

\newtheorem{example}{Example}
\newtheorem{proposition}[theorem]{Proposition}
\usepackage{graphicx}
\usepackage{subfig}
\usepackage{capt-of}
\usepackage{caption}
\usepackage{subcaption}

\usepackage{hyperref}

\usepackage{resizegather}
\usepackage{algorithm,algorithmic}
\usepackage{wrapfig}
\usepackage{cleveref}

\def \P{\mathbb{P}} 
\def \E {\mathbb{E}}

\title{A Relative Error-Based Evaluation Framework of Heterogeneous Treatment Effect Estimators\thanks{This paper has been submitted to ICLR 2026.}}

\author{
Jiayi Guo$^{1}$, Haoxuan Li$^{1}$, Ye Tian$^{2}$, Peng Wu$^{3}$\thanks{Corresponding author: pengwu@btbu.edu.cn.}\\
$^1$Peking University  \quad
$^2$University of Hong Kong \quad
$^3$Beijing Technology and Business University
}

\begin{document}

\maketitle

\begin{abstract}
While significant progress has been made in heterogeneous treatment effect (HTE) estimation, the evaluation of HTE estimators remains underdeveloped. In this article, we propose a robust evaluation framework based on relative error, which quantifies performance differences between two HTE estimators. We first derive the key theoretical conditions on the nuisance parameters that are necessary to achieve a robust estimator of relative error. Building on these conditions, we introduce novel loss functions and design a neural network architecture to estimate nuisance parameters and obtain robust estimation of relative error, thereby achieving reliable evaluation of HTE estimators. We provide the large sample properties of the proposed relative error estimator. Furthermore, beyond evaluation, we propose a new learning algorithm for HTE that leverages both the previously HTE estimators and the nuisance parameters learned through our neural network architecture. Extensive experiments demonstrate that our evaluation framework supports reliable comparisons across HTE estimators, and the proposed learning algorithm for HTE exhibits desirable performance. 
\end{abstract}

\section{Introduction}

The estimation of heterogeneous treatment effects (HTEs) has attracted substantial attention across a range of disciplines, including economics~\citep{Imbens-Rubin2015}, marketing~\citep{wager2018estimation}, biology~\citep{Rosenbaum2020}, and medicine~\citep{Hernan-Robins2020}, due to its critical role in understanding individual-level treatment heterogeneity and supporting personalized, context-specific decision-making. 
Various methods have been developed to estimate HTEs; see \citet{Kunzel-etal2019, Caron-etal2022} for comprehensive reviews. Despite their growing popularity, the evaluation and comparison of HTE estimators remain relatively underexplored~\citep{Gao2024}. Assessing estimator performance is crucial in real-world applications, as a reliable evaluation framework can identify the most suitable methods~\citep{Curth-etal2023}, directly impacting downstream tasks. 

Evaluating HTEs is inherently challenging, as the ground truth is not available: only one potential outcome is observed for each  individual, while HTEs are defined as the difference between two. 
To address this, 
researchers often rely on stringent model assumptions~\citep{Saito-etal2020, Mahajan-etal2024} or preprocessing techniques (e.g., matching)~\citep{Rolling-etal2014} to approximate the unobserved counterfactuals, and obtain an estimated treatment effect. 
Our work is motivated by \citet{Gao2024}, who introduced relative error to quantify the performance difference between two estimators, thereby reducing the bias caused by using inaccurately estimated treatment effects as ground truth. 


Despite the significant contributions of \citet{Gao2024}, a notable limitation remains unaddressed. Their estimator
requires that all nuisance parameter estimators (propensity score and outcome regression models) are consistent at a rate faster than $n^{-1/4}$ to achieve consistency and valid confidence intervals for the relative error, which may be too stringent for real-world applications.  
In practice, the outcome regression models for potential outcomes heavily rely on model extrapolation. These models are 
trained separately within the treated and control groups, yet their predictions are applied across the entire dataset.
When there exists a significant distributional difference between the treated and control groups~\citep{pmlr-v125-jeong20a, Qin12062024}, the extrapolated predictions from these models are prone to inaccuracy and bias, potentially leading to unreliable conclusions.
 Therefore, it is desirable to develop methods that reduce reliance on such extrapolation to ensure more robust and trustworthy evaluations.

To address this limitation, we propose a reliable evaluation approach for HTE estimation that retains the desirable properties of the method in~\citet{Gao2024}, while relaxing the requirement for consistent outcome regression models. We show that the proposed estimator of relative error is $\sqrt{n}$-consistent, asymptotically normal, and yields valid confidence intervals, provided that the propensity score model is consistent at a rate faster than $n^{-1/4}$, even if the outcome regression model is inconsistent. 

This robustness is achieved by carefully exploring the relationships between nuisance parameter models. We first derive the key conditions necessary for robustness and then design a novel loss function for estimating outcome regression models. Moreover, since the proposed method still requires a consistent propensity score model, we introduce novel balance regularizers to mitigate this reliance by encouraging the learned propensity scores to satisfy the balance property~\citep{Imai2014}, i.e., ensuring that the expectation of measurable functions of covariates, weighted by the inverse propensity scores, are equal between treated and control groups.  
Furthermore, by combining the novel loss function with balance regularizers, we design a new neural network architecture that more accurately estimates outcome regression and propensity score models, enabling more reliable relative error estimation and, in turn, more robust HTE evaluations.  
The main contributions are summarized as follows.
\begin{itemize}[leftmargin=*]
    \item We reveal the limitations of existing methods and, through theoretical analysis, derive key conditions for estimating the relative error that mitigate these limitations. 
    
    \item 
We propose a reliable HTE evaluation method by designing novel loss functions and introducing a new neural network, enabling more robust estimation of relative error. 

    \item  We conduct extensive experiments to demonstrate the effectiveness of the proposed method. 
\end{itemize}


\section{Preliminaries}

\subsection{Problem Setting}
We introduce notations to formulate the problem of interest. For each individual $i$, let $A_i \in \mathcal{A} = \{0, 1\}$ denote the binary treatment variable, where $A_i=1$ and $A_i=0$ denote  treatment and control. Let $X_i \in \mathcal{X}\subset\mathbb{R}^d$ be the pre-treatment covariates, and $Y_i \in \mathbb{R}$ be the outcome. We adopt the potential outcome framework in causal inference \citep{Rubin1974, Neyman1990}, defining $Y_i(0)$ and $Y_i(1)$ as the potential outcomes under $A_i=0$ and $A_i=1$, respectively. Since each individual receives either the treatment or the control, the observed outcome $Y_i$ satisfies $Y_i=A_i Y_i(1)+(1-A_i)Y_i(0)$.  

The individual treatment effect (ITE) is defined as $Y_i(1)-Y_i(0)$, which represents the treatment effect for a specific individual $i$. However, since only one of $(Y_i(0), Y_i(1))$ is observable, ITE is not identifiable without imposing strong assumptions~\citep{Hernan-Robins2020, pearl2009causality}.  
In practice, the conditional average treatment effect (CATE) is often used to characterize ``individual" treatment effects, defined by 
\[ \tau(x)=\mathbb{E}[Y_i(1)-Y_i(0)|X_i=x], \]
which captures how treatment effects vary across individuals with different covariate values.   
\begin{assumption}[Strongly Ignorability, \citep{rosenbaum1983central}] \label{assump1}
(i) $(Y_i(0), Y_i(1)) \mid X_i$; (ii)  $0 < e(x) \triangleq \P(A_i=1\mid X_i=x) < 1$ for all $x \in \mathcal{X}$, where 
$e(x)$ is the propensity score.  
\end{assumption}

Under the standard strong ignorability assumption, CATE is identified as $\mu_1(x) - \mu_0(x)$ where $\mu_a(x) = \E[ Y_i \mid X_i=x, A_i=a ]$ for $a = 0, 1$ are the outcome regression functions, and various methods have been developed for estimating CATE~\citep{Wager03072018, pmlr-v70-shalit17a}. 
Suppose we have a set of candidate CATE estimators trained on a training set, denoted by $\{\hat \tau_1(x), \cdots,  \hat \tau_K(x)     \}.$  We aim to select the estimator with the highest accuracy in a test dataset $\{(X_i,A_i,Y_i), i=1,\dots, n\}$, which is of size \(n\) and sampled from the super-population  $\mathbb{P}$, and is independent of the training set.


\subsection{Evaluation Metrics: Absolute Error and Relative Error}

For a given estimator $\hat \tau(x)$, its accuracy is typically evaluated using the MSE defined by 
              \[     \phi(\hat \tau)  \triangleq  \E[  ( \hat \tau(X) - \tau(X) )^2     ].           \]
For any two estimators $\hat \tau_1(x)$ and $\hat \tau_2(x)$, the difference in their MSE is 
        \begin{align*}
          \delta(\hat \tau_1, \hat \tau_2) 
          \triangleq {}&    \phi(\hat \tau_1)  -   \phi(\hat \tau_2) = \E[   \hat \tau_1^2(X) - \hat \tau_2^2(X)  - 2 ( \hat \tau_1(X) - \hat \tau_2(X) ) \tau(X)   ].
        \end{align*} 
\citet{Gao2024} refers to $\phi(\hat \tau)$ and $\delta(\hat \tau_1, \hat \tau_2) $ as the absolute error and relative error, respectively. In practice, absolute error is used much more frequently than relative error.  However, \citet{Gao2024} demonstrated that using relative error as the evaluation metric is superior to using absolute error, both theoretically and experimentally, see Section \ref{sec:motivation} for more details. Intuitively, one can see that the key advantage of using relative error over absolute error lies in that it only relies on the first-order term of the unobserved $\tau$, which reduces the impact of estimation error in $\tau$.

Several studies~\citep{pmlr-v67-gutierrez17a, powers2017methodsheterogeneoustreatmenteffect} have used $\E[  ( Y(1) - Y(0) -  \hat \tau(X)  )^2     ]$ 
to evaluate the estimator $\hat \tau(x)$. However, its estimator requires knowing the values of $(Y(0), Y(1))$ that are not observable in real-world applications.   
We note that 
 \[  \E[  ( Y(1) - Y(0) -  \hat \tau(X)  )^2  =\E[  ( \hat \tau(X) - \tau(X) )^2  + \E[ \text{Var}(Y(1)-Y(0) |X ) ], \]
where the second term on the right-hand side is independent of $\hat \tau(x)$. Thus, this metric is essentially equivalent to the absolute error  $  \phi(\hat \tau)$, and we will not discuss it further. For clarity, we provide a notation summary tab, but due to limited space, we present it in Appendix~\ref{app-notation}.


\section{Motivation}  \label{sec:motivation}

In this section, we briefly discuss the advantages of using relative error over absolute error, 
and then analyze the limitations of the method in \citet{Gao2024}, which motivate this work.

The key theoretical advantage of relative error over absolute error is demonstrated through its semiparametric efficient estimators.  
 A semiparametric efficient estimator is considered optimal (or gold standard) in the sense that it has the smallest asymptotic variance under regularity conditions ~\citep{Newey1990, vdv-1998} given the observed test data.  Let $\{\tilde e(x), \tilde \mu_1(x), \tilde \mu_0(x)\}$ 
be the estimators of $\{ e(x), \mu_1(x), \mu_0(x)\}$, which are the nuisance functions to construct semiparametric efficient estimators of absolute error and relative error. Denote $\tilde \tau(x) = \tilde \mu_1(x) - \tilde \mu_0(x)$.   


{\bf Absolute Error}. Given $\hat \tau(x)$, an estimator of $\phi(\hat \tau)$  is constructed as
{\small\begin{align*} 
           \hat \phi(\hat \tau) ={}& \frac 1 n \sum_{i=1}^n \{    \tilde \tau(X_i) - \hat \tau(X_i)  \}^2 +  2 ( \tilde \tau(X_i)  -  \hat \tau(X_i) ) \left ( \frac{ A_i (Y_i - \tilde \mu_1(X_i))  }{ \tilde e(X_i) } - \frac{ (1-A_i) (Y_i - \tilde \mu_0(X_i))  }{ 1 - \tilde e(X_i) }    \right ). 
        \end{align*}}Under Assumption \ref{assump1}, $\hat \phi(\hat \tau)$ is $\sqrt{n}$-consistent, asymptotically normal, and semiparametric efficient, provided that the estimated nuisance parameter satisfy the key Condition \ref{cond1}. 

\begin{condition} \label{cond1}
 $\E[ (\tilde e(X) - e(X) )^2 ] = o_{\P}(n^{-1/2})$, $\E[ (\tilde \mu_a(X) - \mu_a(X) )^2 ] = o_{\P}(n^{-1/2})$ for $a = 0, 1$.   
\end{condition}


{\bf Relative Error}. Likewise, we can construct the estimator of  $\delta(\hat \tau_1, \hat \tau_2) $ given as  
       \begin{align*}
         &\hat \delta(\hat \tau_1, \hat \tau_2)={} \frac 1 n \sum_{i=1}^n  \varphi(Z_i; \tilde \mu_0, \tilde \mu_1, \tilde e), \; \text{where} \; Z_i \triangleq (A_i, X_i, Y_i),
        \end{align*}
 {\small\begin{align*}
   \varphi(Z_i; \tilde \mu_0, \tilde \mu_1, \tilde e)   \triangleq{}& 
\{\hat \tau_1^2(X_i)  - \hat \tau_2^2(X_i)  \}  - 
 2 ( \hat \tau_1(X_i)  -\hat \tau_2(X_i)  ) \cdot \\ &  \left ( \frac{ A_i (Y_i - \tilde \mu_1(X_i))  }{ \tilde e(X_i) } + \tilde \mu_1(X_i)  - \frac{ (1-A_i) (Y_i - \tilde \mu_0(X_i))  }{ 1 - \tilde e(X_i) } -  \tilde \mu_0(X_i)    \right ).
\end{align*}}Under Assumption \ref{assump1}, the $\sqrt{n}$-consistency, asymptotic normality, and semiparametric efficiency of $ \hat \delta(\hat \tau_1, \tau_2)$ rely on the key Condition \ref{cond2} below.  

\begin{condition} \label{cond2}
     $\E[ | \tilde \mu_a(X) - \mu_a(X)| | \tilde e(X) - e(X) | ] = o_{\P}(n^{-1/2})$.
\end{condition}

Condition \ref{cond2} is strictly weaker than Condition \ref{cond1}. Moreover, the estimator $\hat \delta(\hat \tau_1, \hat \tau_2)$  offers several additional advantages over $\hat \phi(\hat \tau)$, see Appendix \ref{app-A} for more details.


{\bf Motivation.} Although $\hat \delta(\hat \tau_1, \hat \tau_2)$  has several desirable properties, a notable limitation is that Condition \ref{cond2} requires all nuisance parameter estimators to be consistent (as $\tilde e(x)$ and $\tilde \mu_a(x)$  generally converge at most at the rate of $n^{-1/2}$), which may be too stringent for real-world applications. 
In practice, the outcome regression model $\tilde \mu_a(x)$ is learned from the data with $A=a$ and then applied to the entire data. It heavily relies on model extrapolation, as there is often a significant distributional difference between the data with $A=a$ and $A=1-a$~\citep{pmlr-v125-jeong20a, Qin12062024}.  
As a result,  $\tilde \mu_a(x)$ is likely to be inaccurate and biased, violating Assumption \ref{cond2}. Therefore, it is beneficial and practical to develop methods that rely less on model extrapolation.  In contrast, the estimation of the propensity score does not depend on extrapolation, making it less susceptible to this issue. 





A natural and practical question arises: Can we develop a method for estimating $\delta(\hat \tau_1, \hat \tau_2)$, that retains all the desirable properties of $\hat \delta(\hat \tau_1, \hat \tau_2)$, while allowing for bias in $\tilde \mu_a(x)$ (relaxing Condition \ref{cond2})?  In this article, we show that this is achievable by carefully exploiting the connection between the propensity score and outcome regression models, and by designing appropriate loss functions. 

\section{Proposed Method}
\label{sec:4}
In this section, we propose a novel method for estimating the relative error $\delta(\hat \tau_1,  \hat \tau_2)$ that retains the desirable properties of $\hat \delta(\hat \tau_1,  \hat \tau_2)$ while simultaneously being robust to bias in $\tilde \mu_a(x)$ for $a = 0, 1$. 
We consider the following working models for the propensity score and outcome regression functions,  
        \begin{align}
                  e(X) = \P( A=1 \mid X  )  ={}&  e( \Phi(X), \gamma )  = \frac{  \exp(  \Phi(X)^\intercal \gamma   ) }{  1 + \exp(  \Phi(X)^\intercal \gamma   )  },  \label{eq1}  \\
                   \mu_a(X) = \E( Y \mid X, A=a  )  ={}&  \mu_a( \Phi(X), \beta_a )  =   \Phi(X)^\intercal \beta_a, \quad a = 0, 1,  \label{eq2}
        \end{align}
 where $\Phi(X)$ is the a representation of $X$. 

To quantify the bias of $\tilde \mu_a(x)$, it is crucial to distinguish between the working model and the true model. We say a working model is misspecified if the true model does not belong to the working model class, and it is correctly specified if the true model is within the working model class. Example \ref{ex1} provides a misspecified example. 

\begin{example}[A misspecified model] \label{ex1}
Consider $X$ as a scalar and assume that the true model is $\mu_a(X) = \E( Y | X, A=a) = X^2 \beta_a^*$, which represents the true data-generating mechanism of $Y$ given $(X, A=a)$. However, if we learn $\E(Y|X, A=a)$ using a linear model, i.e. $\mu_a(X, \beta_a) := X \beta_a, \beta_a \in \mathbb{R}$, we introduce an inductive bias, meaning we can never reach the true value of \(\beta^*\), even though the estimator may converge.  
Specifically, denote $\hat \beta_a$ as the least-square estimator of $\beta_a$. 
By the property of least-square estimator, it converges to $\bar \beta_a := \E[XX]^{-1} \E[XY]$, regardless of whether \(\mu_a(X, \beta_a)\) is correctly specified or not.
Since $\mu_a(X, \beta_a)$ is misspecified, $\bar \beta_a  \neq \beta_a^*$.  
\end{example}

For models \eqref{eq1} and \eqref{eq2}, let $\check \gamma$ and $\check \beta_a$ denote the estimators of $\gamma$ and $\beta_a$, respectively. Define $\bar \gamma$ and $\bar \beta_a$ as the probability limits of $\check \gamma$ and $\check \beta_a$, and denote $\bar e(X) = e( \Phi(X), \bar \gamma )$ and $\bar \mu_a(X) = \mu_a( \Phi(X), \bar \beta_a )$.   
If model \eqref{eq1} is specified correctly, $e(X) = \bar e(X)$; otherwise, $e(X) \neq \bar e(X)$ and their difference represents the systematic bias induced by model misspecification. 
 Similarly, if model \eqref{eq2} is correctly specified, $\bar \mu_a(X) = \mu_a(X)$; otherwise, $\bar \mu_a(X) \neq \mu_a(X)$.  
\emph{It is important to note that $(\check \gamma, \check \beta_0, \check \beta_1)$ always converges to $(\bar \gamma, \bar \beta_0, \bar \beta_1)$, regardless of whether models \eqref{eq1} and \eqref{eq2} are correctly specified.} 

\subsection{Basic Idea}  \label{sec4-1}
Before delving into the details, we outline its basic idea to provide an intuitive understanding.
 
{\bf First}, to retain the semiparametric efficiency, the proposed estimator $\delta(\hat \tau_1, \hat \tau_2)$ preserves the same form of $\hat \delta(\hat \tau_1, \tau_2)$, which is given as 
\begin{align*}
    & \check \delta(\hat \tau_1, \hat \tau_2; \check \gamma, \check \beta_0, \check \beta_1 )={} \frac 1 n \sum_{i=1}^n
    \varphi(Z_i; \check \mu_0, \check \mu_1, \check e),
\end{align*}
where $\check e(X) = e( \Phi(X), \check \gamma)$ and $\check \mu_a(X) = \mu_a( \Phi(X), \check \beta_a )$ for $a = 0, 1$. 
Although $\check \delta(\hat \tau_1, \hat \tau_2; \check \gamma, \check \beta_0, \check \beta_1 )$ and $\delta(\hat \tau_1, \hat \tau_2)$ share the same form, they differ significantly in how they estimate nuisance parameters, which results in the robustness to biases in $\check \mu_a(X)$.

{\bf Second}, we analyze the key conditions necessary to achieve robustness to biases in $\check \mu_a(X)$.  
  By a Taylor expansion of $\check \delta(\hat \tau_1, \hat \tau_2; \check \gamma, \check \beta_0, \check \beta_1 )$ with respect to $(\bar \gamma,  \bar \beta_0, \bar \beta_1)$, we have that 
\begin{align*}
    \check \delta(\hat \tau_1, \hat \tau_2; \check \gamma, \check \beta_0, \check \beta_1 )  & - \check \delta(\hat \tau_1, \hat \tau_2; \bar \gamma, \bar \beta_0, \bar \beta_1 ) = \Delta_\gamma^\intercal   ( \check \gamma - \bar \gamma) +  \Delta_{\beta_0}^\intercal (\check \beta_0 - \bar \beta_0)  + \Delta_{\beta_1}^\intercal (\check \beta_1 - \bar \beta_1) \\   
    {}& + O_{\P}( (\check \gamma - \bar \gamma)^2 +  (\check \gamma - \bar \gamma) ( \check \beta_1 - \bar \beta_1) +  (\check \gamma - \bar \gamma) ( \check \beta_0 - \bar \beta_0)   ),   
\end{align*}
where 
{\small\begin{align*}
   \Delta_\gamma  ={}& - \frac 1 n \sum_{i=1}^n  2 ( \hat \tau_1(X_i) - \hat \tau_2(X_i)  )  \Big ( \frac{ A_i (1 - \bar e(X_i)) (Y_i - \bar \mu_1(X_i)) }{ \bar e(X_i) } +   \frac{ (1-A_i) \bar e(X_i)  (Y_i - \bar \mu_0(X_i))  }{ 1 - \bar e(X_i) } \Big )  \Phi(X_i), \\
    \Delta_{\beta_0} ={}&  -   \frac 1 n \sum_{i=1}^n   2 ( \hat \tau_1(X_i)  -\hat \tau_2(X_i)  ) \left ( 1  - \frac{1- A_i  }{1- \bar e(X_i) } \right )  \Phi(X_i), \\
       \Delta_{\beta_1} ={}&   \frac 1 n \sum_{i=1}^n   2 ( \hat \tau_1(X_i)  -\hat \tau_2(X_i)  )  \left ( 1 -  \frac{ A_i  }{ \bar e(X_i) }  \right )  \Phi(X_i). 
\end{align*}}

Under mild conditions (see Theorem \ref{thm1}),  
the last term of above Taylor expansion is $o_{\P}(n^{-1/2} )$. We note that $\check \delta(\hat \tau_1, \hat \tau_2; \bar \gamma, \bar \beta_0, \bar \beta_1 )$ is a $\sqrt{n}$-consistent and asymptotically normal estimator of $\delta(\hat \tau_1, \hat \tau_2)$ if either $\check e(x)$ is correctly specified, or $(\check \mu_0(x), \check \mu_1(x))$ is correctly specified. Thus, it is robust to biases in $\check \mu_a(x)$ for $a = 0, 1$ and is the ideal estimator we aim to obtain.  
To ensure that $ \check \delta(\hat \tau_1, \hat \tau_2; \check \gamma, \check \beta_0, \check \beta_1 ) $ has the same asymptotic properties as $\check \delta(\hat \tau_1, \hat \tau_2; \bar \gamma, \bar \beta_0, \bar \beta_1 )$, we require that  
\begin{equation} \label{eq3}
\Delta_\gamma^\intercal ( \check \gamma - \bar \gamma) +  \Delta_{\beta_0}^\intercal (\check \beta_0 - \bar \beta_0)  + \Delta_{\beta_1}^\intercal (\check \beta_1 - \bar \beta_1) = o_{\P}(n^{-1/2}),     
\end{equation}
\emph{even when $(\check \mu_0(x), \check \mu_1(x))$ is misspecified.}  
Note that $(\check \gamma, \check \beta_0, \check \beta_1)$ always converges to $(\bar \gamma, \bar \beta_0, \bar \beta_1)$.  
To satisfy Eq. \eqref{eq3}, it suffices for $\Delta_\gamma$, $\Delta_{\beta_0}$, and $\Delta_{\beta_1}$ to converge to zero at a certain rate. By the central limit theorem, $\Delta_\gamma - \E[\Delta_\gamma] = O_{\P}(n^{-1/2})$, $\Delta_{\beta_0} - \E[\Delta_{\beta_0}] = O_{\P}(n^{-1/2})$, and $\Delta_{\beta_1} -\E[\Delta_{\beta_1} ] = O_{\P}(n^{-1/2})$.  Thus,  Eq. \eqref{eq3} holds provided that 
 \begin{equation*} \label{eq4}
    \E[\Delta_\gamma]  =0, ~ \E[\Delta_{\beta_0}] = 0, ~  \E[\Delta_{\beta_1} ] =0, 
 \end{equation*}
which is equivalent to the following equations: 
    \begin{align} \label{eq5} \begin{cases}
       &  \E \left [  ( \hat \tau_1(X_i) - \hat \tau_2(X_i)  )  \left ( \frac{ A_i (1 - \bar e(X_i)) (Y_i - \bar \mu_1(X_i))  }{ \bar e(X_i) }  +   \frac{ (1-A_i) \bar e(X_i)  (Y_i - \bar \mu_0(X_i))  }{ 1 - \bar e(X_i) }  \right )\Phi(X_i)  \right ]      = 0, \\
        &  \E \left [  ( \hat \tau_1(X_i)  -\hat \tau_2(X_i)  )  \left ( 1 -  \frac{ A_i  }{ \bar e(X_i) }  \right )  \Phi(X_i) \right ]   = 0, \\
         & \E \left [  ( \hat \tau_1(X_i)  -\hat \tau_2(X_i)  ) \left ( 1  - \frac{1- A_i  }{1- \bar e(X_i) } \right )  \Phi(X_i) \right ]    = 0.  
    \end{cases} \end{align}

\subsection{Novel Loss for Nuisance Parameter Estimation}   \label{sec4-2}

To ensure that the first term in Eq. \eqref{eq5} holds, we design the weighted least square loss function for $(\beta_0, \beta_1$) as follows:   
\begin{gather*}
       \mathcal{L}_{\text{wls}}(\beta_0, \beta_1; \check \gamma)  ={}   \frac 1 n \sum_{i=1}^n ( \hat \tau_1(X_i) - \hat \tau_2(X_i)  ) \left [ \frac{ (1-A_i) \check e(X_i)  \{ Y_i -   \Phi(X)^\intercal \beta_0 \}^2  }{ 1 - \check e(X_i) } + \frac{ A_i (1 - \hat e(X_i)) \{Y_i -  \Phi(X)^\intercal \beta_1 \}^2 }{ \hat e(X_i) } \right ] \label{loss-1}. 
  \end{gather*}These loss functions imply that $(\bar \beta_0, \bar \beta_1) \triangleq  \arg \min_{\beta_a} \E[\mathcal{L}_{\text{wls}}(\beta_0, \beta_1; \bar \gamma) ]$.   
By setting $\partial \E[ \mathcal{L}_{\text{wls}}(\beta_0, \beta_1; \bar \gamma)  ] / \partial \beta_0 \big |_{\beta_0 = \bar \beta_0} = 0$ and $\partial \E[ \mathcal{L}_{\text{wls}}(\beta_0, \beta_1; \bar \gamma) ] / \partial \beta_1 \big |_{\beta_1 = \bar \beta_1}= 0$, one can see that the first term in Eq. \eqref{eq5} holds even if $(\check \mu_0(x), \check \mu_1(x))$ is misspecified. 



For learning \(\gamma\), note that Eq.~\eqref{eq5} imposes \(2d\) linear constraints, while \(\gamma \in \mathbb{R}^d\) has only \(d\) degrees of freedom. This makes the system over-constrained. 
To address this, following the soft-margin formulation of support vector machines~\citep{pml1Book}, we introduce slack variables \(\xi, \eta \in \mathbb{R}^d\) to allow controlled constraint violations, and penalize their magnitudes in the objective. 
Formally, we solve:
{\small\begin{align*}
\min_{\gamma, \xi, \eta} \quad & - \frac{1}{n} \sum_{i=1}^n \left[ A_i \log(e(X_i)) + (1 - A_i)\log(1-e(X_i)) \right] + c \sum_{j=1}^{d} (\xi_j + \eta_j)\\
\text{s.t.} \quad & e(X_i) = \frac{\exp(\Phi(X_i)^\top \gamma)}{1 + \exp(\Phi(X_i)^\top \gamma)}, \quad i = 1, \dots, n,\\
& \left| \frac{1}{n} \sum_{i=1}^n (\hat{\tau}_1(X_i) - \hat{\tau}_2(X_i)) \left(1 - \frac{A_i}{e(X_i)}\right) \Phi_j(X_i) \right| \leq \xi_j,\quad \forall j,\\
& \left| \frac{1}{n} \sum_{i=1}^n (\hat{\tau}_1(X_i) - \hat{\tau}_2(X_i)) \left(1 - \frac{1 - A_i}{1 - e(X_i)}\right) \Phi_j(X_i) \right| \leq \eta_j,\quad \forall j,\\
& \xi_j, \eta_j \geq 0,\quad j = 1, \dots, d.
\end{align*}}where \(c\) is a given hyperparameter. 
In practice, we convert the above constrained optimization into two unconstrained loss terms: 
{\small\begin{equation*} \label{eq:ce_loss}
\mathcal{L}_{\text{ce}} = - \frac{1}{n} \sum_{i=1}^n \left[ A_i \log(e(X_i)) + (1 - A_i)\log(1-e(X_i)) \right],
\end{equation*}}
{\small\begin{equation*} \label{eq:const_loss}
\begin{aligned}
\mathcal{L}_{\text{const}} = c \sum_{j=1}^{d} (\xi_j + \eta_j)  + \rho \cdot \left\| 
\begin{bmatrix}
\max\left\{ \left| \frac{1}{n} \sum_{i=1}^n (\hat{\tau}_1(X_i) - \hat{\tau}_2(X_i)) \left(1 - \frac{A_i}{e(X_i)} \right) \Phi(X_i) \right| - \xi,\; 0 \right\} \\
\max\left\{ \left| \frac{1}{n} \sum_{i=1}^n (\hat{\tau}_1(X_i) - \hat{\tau}_2(X_i)) \left(1 - \frac{1 - A_i}{1 - e(X_i)} \right) \Phi(X_i) \right| - \eta,\; 0 \right\} \\
\max(-\xi,\; 0) \\
\max(-\eta,\; 0)
\end{bmatrix}
\right\|_2,
\end{aligned}
\end{equation*}}where \(\rho > 0\) is a penalty parameter encouraging constraint satisfaction.  
\subsection{Constructing Neural Network} \label{sec4-3}

Building on the novel constraint loss introduced in Section~\ref{sec4-2}, we propose a new neural network architecture, 
inspired by the Dragonnet structure~\citep{NEURIPS2019_8fb5f8be}.
 The proposed network takes input features \(x \in \mathbb{R}^d\), and first passes them through multiple fully connected layers to produce the shared representation \(\Phi(x) \in \mathbb{R}^m\). This representation is then fed into three separate heads: a control outcome head \(\mu_0(x)\), predicting the potential outcome under control; a treated outcome head \(\mu_1(x)\), predicting the potential outcome under treatment; a treatment head \(e(x)\), estimating the propensity score via a sigmoid activation.
    
The control outcome head and the treated outcome head contribute to the weighted least square loss \(\mathcal{L}_{\text{wls}}\), while \(\mathcal{L}_{\text{ce}}\) and \(\mathcal{L}_{\text{const}}\) are computed by the treatment head and the shared representation. During training, we minimize the total training loss given by:
\[
\mathcal{L} = \mathcal{L}_{\mathrm{wls}} + \lambda_1 \mathcal{L}_{\mathrm{ce}} + \lambda_2 \mathcal{L}_{\mathrm{const}}.
\]

This formulation encourages the propensity model \(e(X)\) and the outcome model \(\mu_a(X)\) to satisfy Eq.~\eqref{eq5},  providing a reliable estimation that can be used in computing the estimated relative error \(\hat{\delta}\) mentioned in Section~\ref{sec:motivation}. For clarity, we provide a schematic illustration of the network architecture in Appendix~\ref{app-ill}.

\subsection{Theoretical Analysis}  \label{sec4-4}

We analyze the large sample properties of the proposed estimator $\check \delta(\hat \tau_1, \hat \tau_2; \check \gamma, \check \beta_0, \check \beta_1 )$. 

\begin{theorem}\label{thm1}
If the propensity score model is correctly specified, and $\check \gamma$, $\check \beta_0$ as well as $\check \beta_1$ converge to their probability limits at a rate  faster than $n^{-1/4}$, then we have  
\begin{align*}
\sqrt{n} \{\check \delta(\hat \tau_1, \hat \tau_2; \check \gamma, \check \beta_0, \check \beta_1 )  - \delta(\hat \tau_1, \hat \tau_2)\} \xrightarrow{d} \mathcal{N}(0, \sigma^{2}),   
\end{align*}
where $\sigma^{2} =  \text{Var}\{\varphi(Z; \bar u_{0}, \bar u_{1}, \bar e )\}$ and $\xrightarrow{d}$ means convergence in distribution. 
\end{theorem}

Theorem \ref{thm1} shows that the proposed estimator 
 is $\sqrt{n}$-consistent and asymptotically normal. 
 These properties hold even when the outcome regression model is misspecified, as long as $\check \gamma$, $\check \beta_0$, and $\check \beta_1$ converge to their respective probability limits at a rate faster than $n^{-1/4}$. This condition is readily satisfied, as $(\check \gamma, \check \beta_0, \check \beta_1)$ always converge to their probability limits $(\bar \gamma, \bar \beta_0, \bar \beta_1)$, and a variety of flexible machine learning methods can achieve the required convergence rates~\citep{Chernozhukov-etal-2018, Semenova-Chernozhukov}.
 

Based on Theorem \ref{thm1}, we can obtain a valid asymptotic $(1-\eta)$ confidence interval of $\delta(\hat \tau_1, \hat \tau_2)$.  

\begin{proposition}\label{prop}
Under the conditions in Theorem~\ref{thm1},
a consistent estimator of $\sigma^{2}$ is 
\begin{align*}
\hat{\sigma}^{2} = \frac{1}{n} \sum^{n}_{i=1}\left\{\varphi(Z_{i}; \check u_{0}, \check u_{1}, \check e ) -  \check \delta(\hat \tau_1, \hat \tau_2; \check \gamma, \check \beta_0, \check \beta_1 ) \right\}^{2}, 
\end{align*}
an asymptotic $(1-\eta)$ confidence interval for $\delta(\hat \tau_1, \hat \tau_2)$ is $\check \delta(\hat \tau_1, \hat \tau_2; \check \gamma, \check \beta_0, \check \beta_1 ) \pm z_{\eta/2}\sqrt{\hat{\sigma}^2/n}$, where $z_{\eta/2}$ is the $(1-\eta/2)$ quantile of the standard normal distribution. 
\end{proposition}

Proposition \ref{prop} shows that a valid asymptotic confidence interval for $\delta(\hat \tau_1, \hat \tau_2)$ is achievable even with a misspecified outcome model, unlike previous methods that require correct specification.
This further indicates the robustness of the proposed method. 

\section{Enhanced Estimation of Heterogeneous Treatment Effects}
In this section, building on the evaluation framework proposed in Section \ref{sec:4}, we extend the idea to develop a learning method for CATE. 
In general, a reliable evaluation method can naturally serve as a basis for developing a learning method.
In our proposed approach, for any given pair of CATE estimators $\hat \tau_k(x)$ and $\hat \tau_{k'}(x)$, the proposed neural network architecture introduced in Section \ref{sec4-3} can output the corresponding estimates of outcome regression functions. We denote them as $\check \mu_0(x; \hat \tau_k, \hat \tau_{k'})$ and $\check \mu_1(x; \hat \tau_k, \hat \tau_{k'})$, emphasizing their dependence on $\hat \tau_k(x)$ and $\hat \tau_{k'}(x)$. This leads to a new CATE estimator, defined as
        \[    \check \tau(x; \hat \tau_k, \hat \tau_{k'}) = \check \mu_1(x; \hat \tau_k, \hat \tau_{k'}) - \check \mu_0(x; \hat \tau_k, \hat \tau_{k'}).  \]

Clearly, the performance of the estimator heavily depends on the choice of CATE estimators $\check \tau(x; \hat \tau_k, \hat \tau_{k'})$. However, due to the fundamental challenge in evaluating CATE (i.e., the absence of ground truth), it is difficult to develop a direct strategy for selecting them.  
To mitigate this issue, we propose the following aggregation strategy for estimating CATE,  
  \[    \check \tau(x) =  \frac{2}{|\mathcal{K}|(|\mathcal{K}| - 1) } \sum_{k, k' \in \mathcal{K}}  \check \mu_1(x; \hat \tau_k, \hat \tau_{k'}) - \check \mu_0(x; \hat \tau_k, \hat \tau_{k'})   \]
where $\mathcal{K} = \{1, 2, \ldots, K\}$ is the index set for the candidate CATE estimators. This aggregated estimator aims to stabilize and improve the estimation of CATE by averaging over all pairs of candidate estimators. When $K$ is large, averaging over all pairs can be computationally burdensome. In such cases, one can randomly select a subset of pairs $\check \tau(x; \hat \tau_k, \hat \tau_{k'})$ and compute their average instead. 
Surprisingly, our experiments show that this estimator performs exceptionally well, even surpassing the performance of any single candidate estimator. 



\vspace{-8pt}
\section{Experiments}  \label{sec5}
\vspace{-8pt}
\subsection{Experimental Setup}
\vspace{-8pt}
\textbf{Datasets and Processing.} Following previous studies~\citep{yoon2018ganite, NEURIPS2018_a50abba8,louizos2017causal}, we choose one semi-synthetic dataset \textbf{IHDP}, and two real datasets, \textbf{Twins} and \textbf{Jobs}, to conduct our experiments. The \textbf{Twins} dataset is constructed from all twin births in the United States between 1989 and 1991~\citep{10.1093/qje/120.3.1031}, owning 5271 samples with 28 different covariates. 
 The \textbf{IHDP} dataset is used to estimate the effect of specialist home visits on infants’ future cognitive test scores, containing 747 samples (139 treated and 608 control), each with 25 pre-treatment covariates, while the \textbf{Jobs} dataset focuses on estimating the impact of job training programs on individuals’ employment status, including 297 treated units, 425 control units from the experimental sample, and 2490 control units from the observational sample. We provide more dataset details in the Appendix~\ref{app-B1}. We randomly split each dataset into training and test sets in a 2:1 ratio, and repeat the experiments 50 times for the \textbf{Twins}, 100 times for the \textbf{IHDP}, and 20 times for the \textbf{Jobs}.

{\bf Evaluation Metrics.}  
We consider two classes of evaluation metrics below.
\vspace{-6pt} 
\begin{itemize}[leftmargin=*]
 \item
 We assess the proposed relative error estimator using two key metrics: 
(i) the coverage probability of its confidence interval (named coverage rate), and
(ii) the probability of correctly identifying the better estimator (i.e., selecting the true winner, named selection accuracy). 
In practice, we only pick the winner when the confidence interval for the relative error does not contain zero, otherwise, no selection will be made. We calculate the coverage rate of the\textbf{ targeted 90\% }confidence intervals and selection accuracy.
\item 
For evaluating the performance of CATE estimation of our novel network, following previous studies~\citep{pmlr-v70-shalit17a,shi2019adapting, louizos2017causal}, we compute the Precision in Estimation of Heterogeneous Effect (PEHE)~\citep{Hill01012011}, where \(\sqrt{\epsilon_\text{PEHE}} = \sqrt{ \frac{1}{n} \sum_{i=1}^n \left( \hat{\tau}(x_i) - \tau(x_i) \right)^2 }\), and
the absolute error on the ATE, $\epsilon_{\text{ATE}} = | \text{ATE} - \widehat{\text{ATE}}|$ , 
where 
\(
\text{ATE} = \frac{1}{n} \sum_{i=1}^n (y_i^1 - y_j^0),
\)
in which $y_i^1$ and $y_j^0$ are the true potential outcomes.
\end{itemize}

\vspace{-8pt}
\begin{figure}[t]
    \centering
    \begin{minipage}{0.48\textwidth} \
        \centering
        \includegraphics[width=\textwidth]{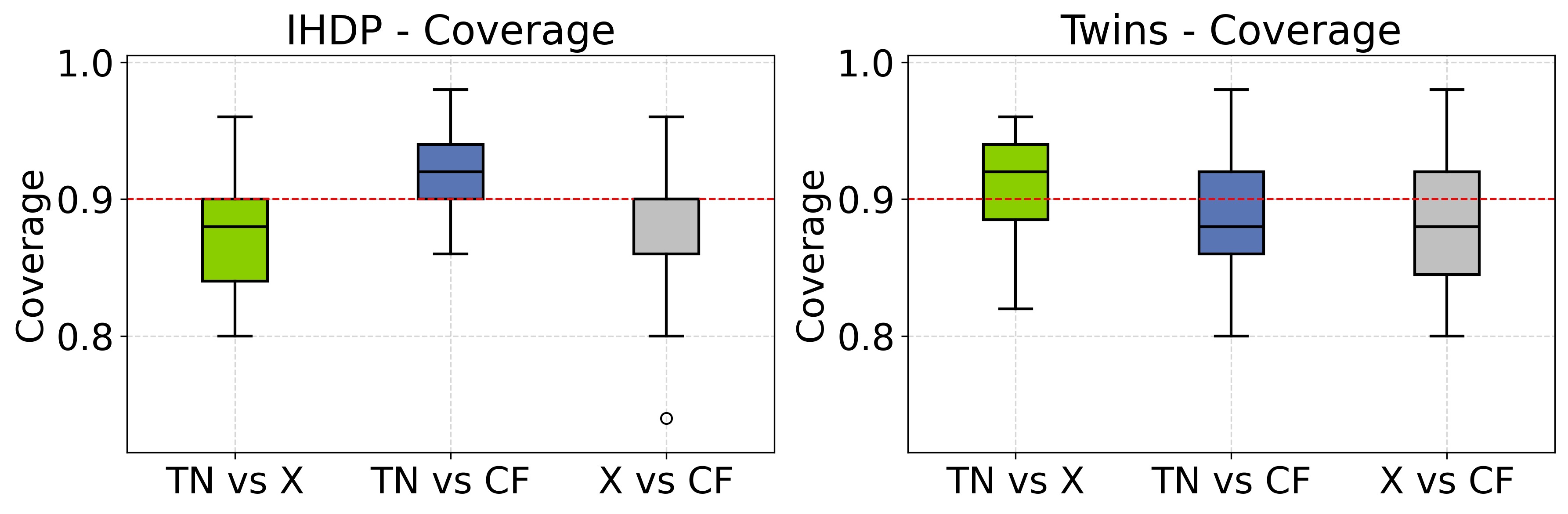}
        \caption{Coverage rate on IHDP and Twins.}
        \label{fig:coverage}
    \end{minipage}
    \hfill
    \begin{minipage}{0.50\textwidth}
        \centering
        \includegraphics[width=\textwidth]{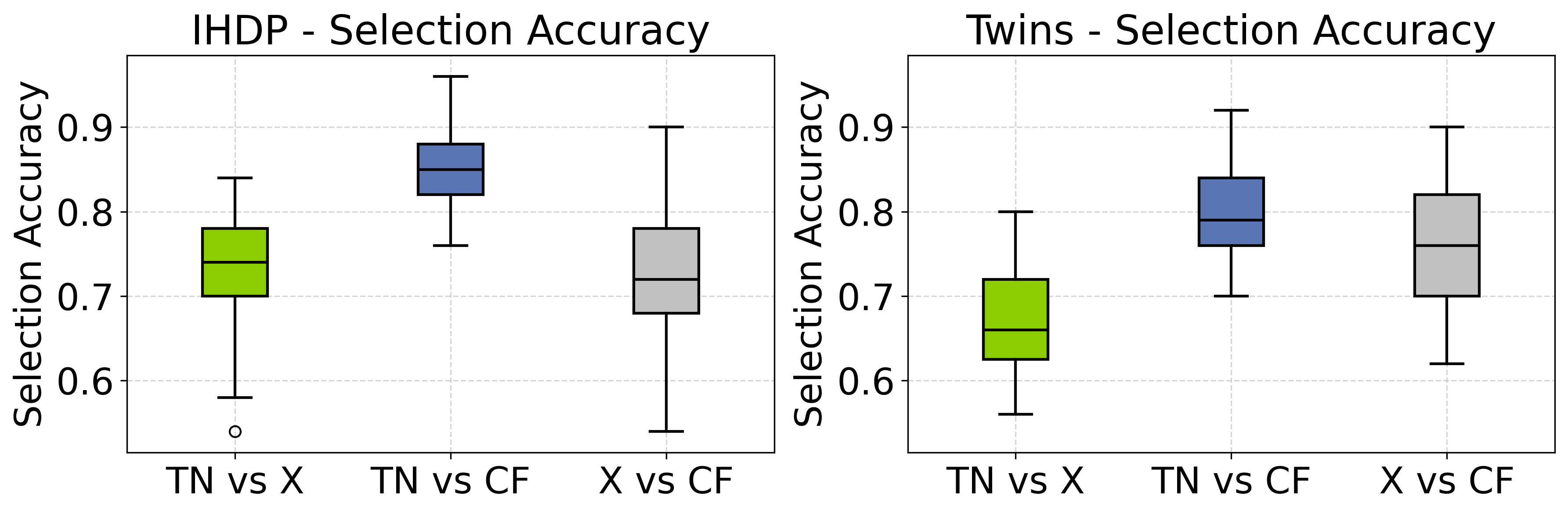}
        \caption{\small Selection accuracy on IHDP and Twins.}
        \label{fig:selection}
    \end{minipage}
    \vspace{-6pt}
\end{figure}

\begin{table}[t]
\centering
\caption{CATE estimation performance on the IHDP and Twins datasets (in-sample and out-of-sample). The best results are bolded.}
\vspace{-8pt}
\label{tab:main}
\resizebox{\textwidth}{!}{
\begin{tabular}{l|cc|cc|cc|cc}
\toprule
\multicolumn{1}{l|}{} &
\multicolumn{4}{c|}{\textbf{IHDP}} & \multicolumn{4}{c}{\textbf{Twins}} \\
 \cmidrule(lr){2-5} \cmidrule(lr){6-9}
\textbf{Method} 
& $\sqrt{\epsilon_{\text{PEHE}}^{\text{in}}}$ & $\epsilon_{\text{ATE}}^{\text{in}}$
& $\sqrt{\epsilon_{\text{PEHE}}^{\text{out}}}$ & $\epsilon_{\text{ATE}}^{\text{out}}$
& $\sqrt{\epsilon_{\text{PEHE}}^{\text{in}}}$ & $\epsilon_{\text{ATE}}^{\text{in}}$
& $\sqrt{\epsilon_{\text{PEHE}}^{\text{out}}}$ & $\epsilon_{\text{ATE}}^{\text{out}}$ \\
\midrule
LinDML     & 1.053 $\pm$ 0.134 & 0.580 $\pm$ 0.152 & 1.085 $\pm$ 0.187 & 0.574 $\pm$ 0.176 & 0.295 $\pm$ 0.005 & 0.013 $\pm$ 0.009 & 0.296 $\pm$ 0.008 & 0.013 $\pm$ 0.010 \\
SpaDML     & 0.832 $\pm$ 0.119 & 0.252 $\pm$ 0.185 & 0.866 $\pm$ 0.112 & 0.280 $\pm$ 0.183 & 0.300 $\pm$ 0.008 & 0.046 $\pm$ 0.030 & 0.303 $\pm$ 0.010 & 0.046 $\pm$ 0.033 \\
CForest    & 0.891 $\pm$ 0.121 & 0.419 $\pm$ 0.182 & 0.903 $\pm$ 0.127 & 0.403 $\pm$ 0.185 & 0.297 $\pm$ 0.005 & 0.012 $\pm$ 0.008 & 0.306 $\pm$ 0.008 & 0.013 $\pm$ 0.011 \\
X-Learner  & 0.971 $\pm$ 0.178 & 0.196 $\pm$ 0.137 & 0.987 $\pm$ 0.196 & 0.207 $\pm$ 0.141 & 0.293 $\pm$ 0.005 & 0.022 $\pm$ 0.014 & 0.294 $\pm$ 0.008 & 0.024 $\pm$ 0.016 \\
S-Learner  & 0.920 $\pm$ 0.102 & 0.212 $\pm$ 0.100 & 0.950 $\pm$ 0.111 & 0.205 $\pm$ 0.117 & 0.298 $\pm$ 0.011 & 0.057 $\pm$ 0.042 & 0.299 $\pm$ 0.010 & 0.059 $\pm$ 0.042 \\
TARNet     & 0.896 $\pm$ 0.054 & 0.279 $\pm$ 0.084 & 0.920 $\pm$ 0.070 & 0.266 $\pm$ 0.117 & 0.292 $\pm$ 0.011 & 0.090 $\pm$ 0.047 & 0.294 $\pm$ 0.019 & 0.091 $\pm$ 0.045 \\
Dragonnet  & 0.840 $\pm$ 0.046 & 0.124 $\pm$ 0.089 & 0.867 $\pm$ 0.087 & 0.134 $\pm$ 0.092 & 0.292 $\pm$ 0.004 & 0.080 $\pm$ 0.008 & 0.290 $\pm$ 0.007 & 0.092 $\pm$ 0.011 \\
DRCFR      & 0.741 $\pm$ 0.068 & 0.186 $\pm$ 0.138 & 0.760 $\pm$ 0.090 & 0.185 $\pm$ 0.135 & 0.290 $\pm$ 0.004 & 0.075 $\pm$ 0.007 & 0.288 $\pm$ 0.007 & 0.076 $\pm$ 0.010 \\
SCIGAN     & 0.898 $\pm$ 0.374 & 0.358 $\pm$ 0.509 & 0.919 $\pm$ 0.369 & 0.358 $\pm$ 0.502 & 0.296 $\pm$ 0.037 & 0.041 $\pm$ 0.044 & 0.293 $\pm$ 0.039 & 0.040 $\pm$ 0.047 \\
DESCN      & 0.793 $\pm$ 0.187 & 0.133 $\pm$ 0.106 & 0.835 $\pm$ 0.197 & 0.140 $\pm$ 0.112 & 0.296 $\pm$ 0.060 & 0.059 $\pm$ 0.043 & 0.293 $\pm$ 0.063 & 0.058 $\pm$ 0.042 \\
ESCFR      & 0.802 $\pm$ 0.041 & 0.111 $\pm$ 0.070 & 0.841 $\pm$ 0.074 & 0.135 $\pm$ 0.076 & 0.290 $\pm$ 0.004 & 0.075 $\pm$ 0.007 & 0.288 $\pm$ 0.007 & 0.076 $\pm$ 0.010 \\
\midrule
\textbf{Ours} & \textbf{0.638 $\pm$ 0.138} & \textbf{0.090 $\pm$ 0.087} & \textbf{0.670 $\pm$ 0.150} & \textbf{0.105 $\pm$ 0.099} & \textbf{0.284 $\pm$ 0.005} & \textbf{0.009 $\pm$ 0.005} & \textbf{0.286 $\pm$ 0.007} & \textbf{0.009 $\pm$ 0.006} \\
\bottomrule
\end{tabular}
}
\vspace{-15pt}
\end{table}

\textbf{Baselines and Experimental Details.} 
To evaluate the performance of relative error estimation, we select three representative estimators from different methodological families: Causal Forest (tree-based)~\citep{athey2019estimatingtreatmenteffectscausal}, X-Learner (meta-learner)~\citep{doi:10.1073/pnas.1804597116}, and TARNet (representation learning)~\citep{ pmlr-v70-shalit17a}. We estimate their pairwise relative errors and evaluate the estimation performances. Although \citeauthor{Gao2024}’s work does not propose a concrete learning method, we follow their choice of nuisance estimators (Linear Regression, Boosting)  to compute relative errors for reference (see Appendix~\ref{app-gao}).

For CATE estimation, the baselines include Causal Forest~\citep{athey2019estimatingtreatmenteffectscausal}, meta-learners (X-Learner, S-Learner)~\citep{doi:10.1073/pnas.1804597116}, double machine learning (Linear DML, Sparse Linear DML)~\citep{chernozhukov2024doubledebiasedmachinelearningtreatment}, TARNet~\citep{pmlr-v70-shalit17a}, Dragonnet~\citep{NEURIPS2019_8fb5f8be}, DR-CFR~\citep{Hassanpour2020Learning}, SCIGAN~\citep{DBLP:journals/corr/abs-2002-12326}, DESCN~\citep{Zhong_2022} and ESCFR~\citep{wang2023optimaltransporttreatmenteffect}. In addition, see Appendix \ref{imp_app} for training details of hyperparameter tuning range.

\vspace{-6pt}
\subsection{Experimental Results}
\vspace{-6pt}
\textbf{Quality of Relative Error Estimation.}  We first  evaluate the performance of relative error estimation, comparing different pairs of HTE estimators.
In Figures~\ref{fig:coverage} and \ref{fig:selection}, we present the coverage of the 90\% confidence intervals  
and the accuracy of selecting the better HTE estimator on the test sets, respectively, where TN stands for TARNet, CF stands for Causal Forest, and X stands for X-Learner, and the red dashed line marks the target level of 90\%. 
From these two figures, our method successfully achieves the target coverage, and 
provide trustworthy advice on the selection  across different pairs of HTE estimators. 
These results demonstrate the validity of our uncertainty quantification and estimator selection.

\textbf{Accuracy of the CATE Estimation.} We then evaluate the performance of CATE estimation learned by our novel network and compare it with competing baselines. 
We average over 100 realizations of our networks in IHDP and 50 realizations in Twins. The results are presented in Table~\ref{tab:main}. Our proposed method achieves the best performance across all metrics, with the lowest $\sqrt{\epsilon_{\text{PEHE}}}$ and $\epsilon_{\text{ATE}}$ on both datasets. This demonstrates its ability to accurately estimate CATE.  In addition, we report results on the \textbf{Jobs} dataset in Appendix~\ref{app-B2} due to limited space.  

\begin{table}[t]
\centering
\caption{Sensitivity analysis on the hyperparameter $\lambda_2$ (weight of constraint loss) for IHDP and Twins datasets. The best hyperparameter values and results are in bold.}
\vspace{-8pt}
\label{tab:sensitivity_lambda2}
\resizebox{\textwidth}{!}{
\begin{tabular}{l|cc|cc|cc|l|cc|cc|cc}
\toprule
\multicolumn{7}{c|}{\textbf{IHDP}} & \multicolumn{7}{c}{\textbf{Twins}} \\
\cmidrule(lr){1-7} \cmidrule(lr){8-14}
\textbf{Value} 
& $\sqrt{\epsilon_{\text{PEHE}}^{\text{in}}}$ & $\epsilon_{\text{ATE}}^{\text{in}}$ 
& $\sqrt{\epsilon_{\text{PEHE}}^{\text{out}}}$ & $\epsilon_{\text{ATE}}^{\text{out}}$ 
& Coverage & Selection 
& \textbf{Value} 
& $\sqrt{\epsilon_{\text{PEHE}}^{\text{in}}}$ & $\epsilon_{\text{ATE}}^{\text{in}}$ 
& $\sqrt{\epsilon_{\text{PEHE}}^{\text{out}}}$ & $\epsilon_{\text{ATE}}^{\text{out}}$ 
& Coverage & Selection \\
\midrule
0.01  & 0.860 & 0.216 & 0.902 & 0.238 & 0.85 & 0.50  & 0.005 & 0.319 & 0.029 & 0.331 & 0.027 & 0.82 & 0.38 \\
0.1   & 0.800 & 0.142 & 0.837 & 0.158 & 0.91 & 0.61  & 0.05  & 0.289 & 0.016 & 0.292 & 0.015 & 0.82 & 0.84 \\
0.5   & 0.714 & 0.099 & 0.747 & 0.118 & 0.95 & 0.78  & 0.25  & 0.297 & 0.018 & 0.297 & 0.020 & 0.86 & 0.42 \\
\textbf{1}     & \textbf{0.638} & \textbf{0.090} & \textbf{0.670} & \textbf{0.105} & \textbf{0.96} & \textbf{0.80}  & \textbf{0.5}   & \textbf{0.284} & \textbf{0.009} & \textbf{0.286} & \textbf{0.009} & \textbf{0.94} & \textbf{0.94} \\
5     & 0.715 & 0.099 & 0.748 & 0.116 & 0.94 & 0.77  & 2.5   & 0.285 & 0.011 & 0.287 & 0.012 & 0.94 & 0.92 \\
10    & 0.795 & 0.157 & 0.830 & 0.172 & 0.90 & 0.60  & 5     & 0.289 & 0.028 & 0.290 & 0.026 & 0.80 & 0.86 \\
100   & 0.801 & 0.156 & 0.836 & 0.170 & 0.90 & 0.60  & 50    & 0.287 & 0.024 & 0.288 & 0.023 & 0.84 & 0.88 \\
\bottomrule
\end{tabular}
}
\vspace{-8pt}
\end{table}

\begin{table}[t]
\centering
\caption{Ablation study results on the IHDP and Twins datasets.}
\vspace{-8pt}
\label{tab:ablation_full}
\resizebox{\textwidth}{!}{
\begin{tabular}{l|cc|cc|cc|cc|cc|cc}
\toprule
\multicolumn{1}{l|}{} &
\multicolumn{6}{c|}{\textbf{IHDP}} & \multicolumn{6}{c}{\textbf{Twins}} \\
\cmidrule(lr){2-7} \cmidrule(lr){8-13}
\textbf{Training Loss} 
& $\sqrt{\epsilon_{\text{PEHE}}^{\text{in}}}$ & $\epsilon_{\text{ATE}}^{\text{in}}$ 
& $\sqrt{\epsilon_{\text{PEHE}}^{\text{out}}}$ & $\epsilon_{\text{ATE}}^{\text{out}}$ 
& Coverage & Selection 
& $\sqrt{\epsilon_{\text{PEHE}}^{\text{in}}}$ & $\epsilon_{\text{ATE}}^{\text{in}}$ 
& $\sqrt{\epsilon_{\text{PEHE}}^{\text{out}}}$ & $\epsilon_{\text{ATE}}^{\text{out}}$ 
& Coverage & Selection \\
\midrule
$\mathcal{L}_{\text{wls}}$ \& $\mathcal{L}_{\text{const}}$                    & 0.725 & 0.101 & 0.758 & 0.122 & 0.92 & 0.71 & 0.284 & 0.013 & 0.287 & 0.013 & 0.94 & 0.92 \\
$\mathcal{L}_{\text{wls}}$ \& $\mathcal{L}_{\text{ce}}$                 & 3.495 & 2.879 & 3.531 & 2.900 & 0.88 & 0.14 & 0.319 & 0.028 & 0.328 & 0.026 & 0.82 & 0.14 \\

\midrule
\textbf{Full (Ours)} & \textbf{0.638} & \textbf{0.090} & \textbf{0.670} & \textbf{0.105} & \textbf{0.96} & \textbf{0.80} & \textbf{0.284} & \textbf{0.009} & \textbf{0.286} & \textbf{0.009} & \textbf{0.94} & \textbf{0.94} \\
\bottomrule
\end{tabular}
}
\vspace{-12pt}
\end{table}

\textbf{Sensitive Analysis.} The hyperparameter \(\lambda_2\) before the constraint loss \(\mathcal{L}_{\text{const}}\), 
\(\lambda_1\) before the cross entropy loss \(\mathcal{L}_{\text{ce}}\) and the penalty weight \(\rho\) in the constraint loss  
play important roles in our training. In order to explore under which parameters our method has the best performance, we conduct sensitivity analysis experiments. We present
the results of \(\lambda_2\) in Table~\ref{tab:sensitivity_lambda2}. 
We observe that both the performance of CATE estimation and the relative error estimation remain relatively stable across a range of $\lambda_2$ values from 0.5 to 5, indicating robustness to this hyperparameter.    
However, when $\lambda_2$ is extremely small (e.g., $\lambda_2 = 0.01$), the performance of the proposed method degrades significantly, indicating the importance of the constraint loss \(\mathcal{L}_{\text{const}}\). 
Also, we perform sensitive analysis for \(\lambda_1\) and \(\rho\), the associated results are provided in the Appendix~\ref{app-B3}.

{\bf Ablation Study.} As shown in Section \ref{sec4-3}, the proposed method involves three loss functions: $\mathcal{L}_{\mathrm{wls}}, \mathcal{L}_{\mathrm{ce}}$, and $\mathcal{L}_{\mathrm{const}}$.
We conduct an ablation study to assess the impact of $\mathcal{L}_{\mathrm{ce}}$ and $\mathcal{L}_{\mathrm{const}}$ on overall performance. The corresponding results are  reported in Table~\ref{tab:ablation_full}. 
Specifically, removing $\mathcal{L}_{\text{const}}$ results in a notable drop in the accuracy of both outcome and relative error estimation, whereas removing $\mathcal{L}_{\text{ce}}$ only causes a moderate decline. These findings highlight the importance of the proposed novel loss $\mathcal{L}_{\text{const}}$, which not only improves HTE estimation accuracy but also facilitates the construction of narrower and more precise confidence intervals for relative error. 
   
\vspace{-12pt}
\section{Conclusion}
\vspace{-8pt}
In this work, we addressed a key challenge in evaluating HTE estimators with less reliance on modeling assumptions for nuisance parameters. 
Building upon the relative error framework, we introduced a novel loss function and balance regularizers that encourage more stable and accurate learning of nuisance parameters. These components were integrated into a new neural network architecture tailored to enhance the reliability of HTE evaluation.  
The proposed evaluation approach retains several desirable statistical properties while relaxing the stringent requirement for consistent outcome regression models, thereby facilitating more reliable comparisons and selection of estimators in real-world applications. A limitation of this work lies in the use of the simple averaging scheme over all estimator pairs for CATE estimation. While this approach improves stability, it may not fully exploit the varying strengths of individual estimators, potentially limiting overall efficiency and precision. Future research is warranted to further address this challenge.



\bibliography{refs.bib}

\begin{thebibliography}{47}
\providecommand{\natexlab}[1]{#1}
\providecommand{\url}[1]{\texttt{#1}}
\expandafter\ifx\csname urlstyle\endcsname\relax
  \providecommand{\doi}[1]{doi: #1}\else
  \providecommand{\doi}{doi: \begingroup \urlstyle{rm}\Url}\fi

\bibitem[{A. Smith} \& {E. Todd}(2005){A. Smith} and {E. Todd}]{ASMITH2005305}
Jeffrey {A. Smith} and Petra {E. Todd}.
\newblock Does matching overcome lalonde's critique of nonexperimental
  estimators?
\newblock \emph{Journal of Econometrics}, 125\penalty0 (1):\penalty0 305--353,
  2005.
\newblock ISSN 0304-4076.
\newblock \doi{https://doi.org/10.1016/j.jeconom.2004.04.011}.
\newblock URL
  \url{https://www.sciencedirect.com/science/article/pii/S030440760400082X}.
\newblock Experimental and non-experimental evaluation of economic policy and
  models.

\bibitem[Almond et~al.(2005)Almond, Chay, and Lee]{10.1093/qje/120.3.1031}
Douglas Almond, Kenneth~Y. Chay, and David~S. Lee.
\newblock The costs of low birth weight*.
\newblock \emph{The Quarterly Journal of Economics}, 120\penalty0 (3):\penalty0
  1031--1083, 08 2005.
\newblock ISSN 0033-5533.
\newblock \doi{10.1093/qje/120.3.1031}.
\newblock URL \url{https://doi.org/10.1093/qje/120.3.1031}.

\bibitem[and(2011)]{Hill01012011}
Jennifer L.~Hill and.
\newblock Bayesian nonparametric modeling for causal inference.
\newblock \emph{Journal of Computational and Graphical Statistics}, 20\penalty0
  (1):\penalty0 217--240, 2011.
\newblock \doi{10.1198/jcgs.2010.08162}.
\newblock URL \url{https://doi.org/10.1198/jcgs.2010.08162}.

\bibitem[Athey \& Wager(2019)Athey and
  Wager]{athey2019estimatingtreatmenteffectscausal}
Susan Athey and Stefan Wager.
\newblock Estimating treatment effects with causal forests: An application,
  2019.
\newblock URL \url{https://arxiv.org/abs/1902.07409}.

\bibitem[Bica et~al.(2020)Bica, Jordon, and van~der
  Schaar]{DBLP:journals/corr/abs-2002-12326}
Ioana Bica, James Jordon, and Mihaela van~der Schaar.
\newblock Estimating the effects of continuous-valued interventions using
  generative adversarial networks.
\newblock \emph{CoRR}, abs/2002.12326, 2020.
\newblock URL \url{https://arxiv.org/abs/2002.12326}.

\bibitem[Caron et~al.(2022)Caron, Baio, and Manolopoulou]{Caron-etal2022}
Alberto Caron, Gianluca Baio, and Ioanna Manolopoulou.
\newblock Estimating individual treatment effects using non-parametric
  regression models: A review.
\newblock \emph{Journal of the Royal Statistical Society: Series A (Statistics
  in Society)}, 185:\penalty0 1115--1149, 2022.

\bibitem[Chernozhukov et~al.(2018)Chernozhukov, Chetverikov, Demirer, Duflo,
  Hansen, Newey, and Robins]{Chernozhukov-etal-2018}
V.~Chernozhukov, D.~Chetverikov, M.~Demirer, E.~Duflo, C.~Hansen, W.~Newey, and
  J.~Robins.
\newblock Double/debiased machine learning for treatment and structural
  parameters.
\newblock \emph{The Econometrics Journal}, 21:\penalty0 1--68, 2018.

\bibitem[Chernozhukov et~al.(2024)Chernozhukov, Chetverikov, Demirer, Duflo,
  Hansen, Newey, and
  Robins]{chernozhukov2024doubledebiasedmachinelearningtreatment}
Victor Chernozhukov, Denis Chetverikov, Mert Demirer, Esther Duflo, Christian
  Hansen, Whitney Newey, and James Robins.
\newblock Double/debiased machine learning for treatment and causal parameters,
  2024.
\newblock URL \url{https://arxiv.org/abs/1608.00060}.

\bibitem[Curth \& Van Der~Schaar(2023)Curth and Van Der~Schaar]{Curth-etal2023}
Alicia Curth and Mihaela Van Der~Schaar.
\newblock In search of insights, not magic bullets: towards demystification of
  the model selection dilemma in heterogeneous treatment effect estimation.
\newblock In \emph{Proceedings of the 40th International Conference on Machine
  Learning}, ICML'23. JMLR, 2023.

\bibitem[Dehejia \& Wahba(2002)Dehejia and Wahba]{10.1162/003465302317331982}
Rajeev~H. Dehejia and Sadek Wahba.
\newblock Propensity score-matching methods for nonexperimental causal studies.
\newblock \emph{The Review of Economics and Statistics}, 84\penalty0
  (1):\penalty0 151--161, 02 2002.
\newblock ISSN 0034-6535.
\newblock \doi{10.1162/003465302317331982}.
\newblock URL \url{https://doi.org/10.1162/003465302317331982}.

\bibitem[Dorie(2016)]{dorie2016npci}
Vincent Dorie.
\newblock vdorie/npci, 2016.
\newblock URL \url{https://github.com/vdorie/npci}.
\newblock GitHub repository.

\bibitem[Gao(2025)]{Gao2024}
Zijun Gao.
\newblock Trustworthy assessment of heterogeneous treatment effect estimator
  via analysis of relative error.
\newblock In \emph{The 28th International Conference on Artificial Intelligence
  and Statistics}, 2025.
\newblock URL \url{https://openreview.net/forum?id=kOTUgBknsK}.

\bibitem[Gutierrez \& Gérardy(2017)Gutierrez and
  Gérardy]{pmlr-v67-gutierrez17a}
Pierre Gutierrez and Jean-Yves Gérardy.
\newblock Causal inference and uplift modelling: A review of the literature.
\newblock In Claire Hardgrove, Louis Dorard, Keiran Thompson, and Florian
  Douetteau (eds.), \emph{Proceedings of The 3rd International Conference on
  Predictive Applications and APIs}, volume~67 of \emph{Proceedings of Machine
  Learning Research}, pp.\  1--13. PMLR, 11--12 Oct 2017.
\newblock URL \url{https://proceedings.mlr.press/v67/gutierrez17a.html}.

\bibitem[Hassanpour \& Greiner(2020)Hassanpour and
  Greiner]{Hassanpour2020Learning}
Negar Hassanpour and Russell Greiner.
\newblock Learning disentangled representations for counterfactual regression.
\newblock In \emph{International Conference on Learning Representations}, 2020.
\newblock URL \url{https://openreview.net/forum?id=HkxBJT4YvB}.

\bibitem[Hern{\'a}n \& Robins(2020)Hern{\'a}n and Robins]{Hernan-Robins2020}
M.A. Hern{\'a}n and J.~M. Robins.
\newblock \emph{Causal Inference: What If}.
\newblock Boca Raton: Chapman and Hall/CRC, 2020.

\bibitem[Imai \& Ratkovic(2014)Imai and Ratkovic]{Imai2014}
Kosuke Imai and Marc Ratkovic.
\newblock Covariate balancing propensity score.
\newblock \emph{Journal of the Royal Statistical Society (Series B)},
  76\penalty0 (1):\penalty0 243--263, 2014.

\bibitem[Imbens \& Rubin(2015)Imbens and Rubin]{Imbens-Rubin2015}
G.~W. Imbens and D.~B. Rubin.
\newblock \emph{Causal Inference For Statistics Social and Biomedical Science}.
\newblock Cambridge University Press, 2015.

\bibitem[Jeong \& Namkoong(2020)Jeong and Namkoong]{pmlr-v125-jeong20a}
Sookyo Jeong and Hongseok Namkoong.
\newblock Robust causal inference under covariate shift via worst-case
  subpopulation treatment effects.
\newblock In Jacob Abernethy and Shivani Agarwal (eds.), \emph{Proceedings of
  Thirty Third Conference on Learning Theory}, volume 125 of \emph{Proceedings
  of Machine Learning Research}, pp.\  2079--2084. PMLR, 09--12 Jul 2020.
\newblock URL \url{https://proceedings.mlr.press/v125/jeong20a.html}.

\bibitem[Jing~Qin \& Huang(2024)Jing~Qin and Huang]{Qin12062024}
Moming~Li Jing~Qin, Yukun~Liu and Chiung-Yu Huang.
\newblock Distribution-free prediction intervals under covariate shift, with an
  application to causal inference.
\newblock \emph{Journal of the American Statistical Association}, 0\penalty0
  (0):\penalty0 1--2, 2024.
\newblock \doi{10.1080/01621459.2024.2356886}.
\newblock URL \url{https://doi.org/10.1080/01621459.2024.2356886}.

\bibitem[Kunzel et~al.(2019)Kunzel, Sekhon, Bickel, and Y]{Kunzel-etal2019}
Soren~R Kunzel, Jasjeet~S Sekhon, Peter~J Bickel, and Bin Y.
\newblock Metalearners for estimating heterogeneous treatment effects using
  machine learning.
\newblock \emph{PNAS}, 116:\penalty0 4156–4165, 2019.

\bibitem[Künzel et~al.(2019)Künzel, Sekhon, Bickel, and
  Yu]{doi:10.1073/pnas.1804597116}
Sören~R. Künzel, Jasjeet~S. Sekhon, Peter~J. Bickel, and Bin Yu.
\newblock Metalearners for estimating heterogeneous treatment effects using
  machine learning.
\newblock \emph{Proceedings of the National Academy of Sciences}, 116\penalty0
  (10):\penalty0 4156--4165, 2019.
\newblock \doi{10.1073/pnas.1804597116}.
\newblock URL \url{https://www.pnas.org/doi/abs/10.1073/pnas.1804597116}.

\bibitem[LaLonde(1986)]{9c8bcb59-753a-33cb-871e-17f05b11d793}
Robert~J. LaLonde.
\newblock Evaluating the econometric evaluations of training programs with
  experimental data.
\newblock \emph{The American Economic Review}, 76\penalty0 (4):\penalty0
  604--620, 1986.
\newblock ISSN 00028282.
\newblock URL \url{http://www.jstor.org/stable/1806062}.

\bibitem[Louizos et~al.(2017)Louizos, Shalit, Mooij, Sontag, Zemel, and
  Welling]{louizos2017causal}
Christos Louizos, Uri Shalit, Joris~M Mooij, David Sontag, Richard Zemel, and
  Max Welling.
\newblock Causal effect inference with deep latent-variable models.
\newblock \emph{Advances in neural information processing systems}, 30, 2017.

\bibitem[Mahajan et~al.(2024)Mahajan, Mitliagkas, Neal, and
  Syrgkanis]{Mahajan-etal2024}
Divyat Mahajan, Ioannis Mitliagkas, Brady Neal, and Vasilis Syrgkanis.
\newblock Empirical analysis of model selection for heterogeneous causal effect
  estimation.
\newblock \emph{arXiv preprint arXiv:2211.01939}, 2024.

\bibitem[Murphy(2022)]{pml1Book}
Kevin~P. Murphy.
\newblock \emph{Probabilistic Machine Learning: An introduction}.
\newblock MIT Press, 2022.
\newblock URL \url{http://probml.github.io/book1}.

\bibitem[Newey(1990)]{Newey1990}
Whitney~K. Newey.
\newblock Semiparametric efficiency bounds.
\newblock \emph{Journal of Applied Econometrics}, 5:\penalty0 99–135, 1990.

\bibitem[Neyman(1990)]{Neyman1990}
Jerzy~Splawa Neyman.
\newblock On the application of probability theory to agricultural experiments.
  essay on principles. section 9.
\newblock \emph{Statistical Science}, 5:\penalty0 465--472, 1990.

\bibitem[Pearl(2009)]{pearl2009causality}
Judea Pearl.
\newblock \emph{Causality}.
\newblock Cambridge university press, 2009.

\bibitem[Powers et~al.(2017)Powers, Qian, Jung, Schuler, Shah, Hastie, and
  Tibshirani]{powers2017methodsheterogeneoustreatmenteffect}
Scott Powers, Junyang Qian, Kenneth Jung, Alejandro Schuler, Nigam~H. Shah,
  Trevor Hastie, and Robert Tibshirani.
\newblock Some methods for heterogeneous treatment effect estimation in
  high-dimensions, 2017.
\newblock URL \url{https://arxiv.org/abs/1707.00102}.

\bibitem[Rolling \& Yang(2014)Rolling and Yang]{Rolling-etal2014}
Craig~A. Rolling and Yuhong Yang.
\newblock Model selection for estimating treatment effects.
\newblock \emph{Journal of the Royal Statistical Society Series B: Statistical
  Methodology}, 76\penalty0 (4):\penalty0 749--769, 2014.

\bibitem[Rosenbaum(2020)]{Rosenbaum2020}
Paul~R. Rosenbaum.
\newblock \emph{Design of Observational Studies}.
\newblock Springer Nature Switzerland AG, second edition, 2020.

\bibitem[Rosenbaum \& Rubin(1983)Rosenbaum and Rubin]{rosenbaum1983central}
Paul~R Rosenbaum and Donald~B Rubin.
\newblock The central role of the propensity score in observational studies for
  causal effects.
\newblock \emph{Biometrika}, 70\penalty0 (1):\penalty0 41--55, 1983.

\bibitem[Rubin(1974)]{Rubin1974}
D.~B. Rubin.
\newblock Estimating causal effects of treatments in randomized and
  nonrandomized studies.
\newblock \emph{Journal of educational psychology}, 66:\penalty0 688--701,
  1974.

\bibitem[Saito \& YasuiAuthors(2023)Saito and YasuiAuthors]{Saito-etal2020}
Yuta Saito and Shota YasuiAuthors.
\newblock Counterfactual cross-validation: stable model selection procedure for
  causal inference models.
\newblock In \emph{Proceedings of the 37th International Conference on Machine
  Learning}, ICML'20. JMLR, 2023.

\bibitem[Semenova \& Chernozhukov(2021)Semenova and
  Chernozhukov]{Semenova-Chernozhukov}
Vira Semenova and Victor Chernozhukov.
\newblock Debiased machine learning of conditional average treatment effects
  and and other causal functions.
\newblock \emph{The Econometrics Journal}, 24:\penalty0 264--289, 2021.

\bibitem[Shalit et~al.(2017{\natexlab{a}})Shalit, Johansson, and
  Sontag]{pmlr-v70-shalit17a}
Uri Shalit, Fredrik~D. Johansson, and David Sontag.
\newblock Estimating individual treatment effect: generalization bounds and
  algorithms.
\newblock In Doina Precup and Yee~Whye Teh (eds.), \emph{Proceedings of the
  34th International Conference on Machine Learning}, volume~70 of
  \emph{Proceedings of Machine Learning Research}, pp.\  3076--3085. PMLR,
  06--11 Aug 2017{\natexlab{a}}.
\newblock URL \url{https://proceedings.mlr.press/v70/shalit17a.html}.

\bibitem[Shalit et~al.(2017{\natexlab{b}})Shalit, Johansson, and
  Sontag]{shalit2017estimatingindividualtreatmenteffect}
Uri Shalit, Fredrik~D. Johansson, and David Sontag.
\newblock Estimating individual treatment effect: generalization bounds and
  algorithms, 2017{\natexlab{b}}.
\newblock URL \url{https://arxiv.org/abs/1606.03976}.

\bibitem[Shi et~al.(2019{\natexlab{a}})Shi, Blei, and
  Veitch]{NEURIPS2019_8fb5f8be}
Claudia Shi, David Blei, and Victor Veitch.
\newblock Adapting neural networks for the estimation of treatment effects.
\newblock In H.~Wallach, H.~Larochelle, A.~Beygelzimer, F.~d\textquotesingle
  Alch\'{e}-Buc, E.~Fox, and R.~Garnett (eds.), \emph{Advances in Neural
  Information Processing Systems}, volume~32. Curran Associates, Inc.,
  2019{\natexlab{a}}.
\newblock URL
  \url{https://proceedings.neurips.cc/paper_files/paper/2019/file/8fb5f8be2aa9d6c64a04e3ab9f63feee-Paper.pdf}.

\bibitem[Shi et~al.(2019{\natexlab{b}})Shi, Blei, and Veitch]{shi2019adapting}
Claudia Shi, David Blei, and Victor Veitch.
\newblock Adapting neural networks for the estimation of treatment effects.
\newblock \emph{Advances in neural information processing systems}, 32,
  2019{\natexlab{b}}.

\bibitem[van~der Vaart(1998)]{vdv-1998}
Aad~W. van~der Vaart.
\newblock \emph{Asymptotic statistics}.
\newblock Cambridge University Press, 1998.

\bibitem[Wager \& Athey(2018{\natexlab{a}})Wager and Athey]{Wager03072018}
Stefan Wager and Susan Athey.
\newblock Estimation and inference of heterogeneous treatment effects using
  random forests.
\newblock \emph{Journal of the American Statistical Association}, 113\penalty0
  (523):\penalty0 1228--1242, 2018{\natexlab{a}}.
\newblock \doi{10.1080/01621459.2017.1319839}.
\newblock URL \url{https://doi.org/10.1080/01621459.2017.1319839}.

\bibitem[Wager \& Athey(2018{\natexlab{b}})Wager and
  Athey]{wager2018estimation}
Stefan Wager and Susan Athey.
\newblock Estimation and inference of heterogeneous treatment effects using
  random forests.
\newblock \emph{Journal of the American Statistical Association}, 113:\penalty0
  1228--1242, 2018{\natexlab{b}}.

\bibitem[Wang et~al.(2023)Wang, Chen, Fan, Li, Liu, Liu, Dai, Wang, Dong, and
  Tang]{wang2023optimaltransporttreatmenteffect}
Hao Wang, Zhichao Chen, Jiajun Fan, Haoxuan Li, Tianqiao Liu, Weiming Liu,
  Quanyu Dai, Yichao Wang, Zhenhua Dong, and Ruiming Tang.
\newblock Optimal transport for treatment effect estimation, 2023.
\newblock URL \url{https://arxiv.org/abs/2310.18286}.

\bibitem[Wu et~al.(2022)Wu, Yuan, Kuang, Li, Wu, Zhu, Zhuang, and
  Wu]{wu2022learning}
Anpeng Wu, Junkun Yuan, Kun Kuang, Bo~Li, Runze Wu, Qiang Zhu, Yueting Zhuang,
  and Fei Wu.
\newblock Learning decomposed representations for treatment effect estimation.
\newblock \emph{IEEE Transactions on Knowledge and Data Engineering},
  35\penalty0 (5):\penalty0 4989--5001, 2022.

\bibitem[Yao et~al.(2018)Yao, Li, Li, Huai, Gao, and
  Zhang]{NEURIPS2018_a50abba8}
Liuyi Yao, Sheng Li, Yaliang Li, Mengdi Huai, Jing Gao, and Aidong Zhang.
\newblock Representation learning for treatment effect estimation from
  observational data.
\newblock In S.~Bengio, H.~Wallach, H.~Larochelle, K.~Grauman, N.~Cesa-Bianchi,
  and R.~Garnett (eds.), \emph{Advances in Neural Information Processing
  Systems}, volume~31. Curran Associates, Inc., 2018.
\newblock URL
  \url{https://proceedings.neurips.cc/paper_files/paper/2018/file/a50abba8132a77191791390c3eb19fe7-Paper.pdf}.

\bibitem[Yoon et~al.(2018)Yoon, Jordon, and Van Der~Schaar]{yoon2018ganite}
Jinsung Yoon, James Jordon, and Mihaela Van Der~Schaar.
\newblock Ganite: Estimation of individualized treatment effects using
  generative adversarial nets.
\newblock In \emph{International Conference on Learning Representations}, 2018.

\bibitem[Zhong et~al.(2022)Zhong, Xiao, Ren, Liang, Yao, Yang, and
  Cen]{Zhong_2022}
Kailiang Zhong, Fengtong Xiao, Yan Ren, Yaorong Liang, Wenqing Yao, Xiaofeng
  Yang, and Ling Cen.
\newblock Descn: Deep entire space cross networks for individual treatment
  effect estimation.
\newblock In \emph{Proceedings of the 28th ACM SIGKDD Conference on Knowledge
  Discovery and Data Mining}, KDD ’22, pp.\  4612–4620. ACM, August 2022.
\newblock \doi{10.1145/3534678.3539198}.
\newblock URL \url{http://dx.doi.org/10.1145/3534678.3539198}.

\end{thebibliography}
\bibliographystyle{iclr2026_conference}

\newpage 
\appendix

\setcounter{theorem}{0}
\setcounter{equation}{0}
 \renewcommand{\theequation}{A.\arabic{equation}}
\newtheorem{lem}{Lemma}[section]
\newtheorem{thm}{Theorem}[section]
\newtheorem{prop}{Proposition}[section]

\section{Notation Summary} \label{app-notation}
\begin{table}[ht]
\centering
\caption{Notation and their meanings.}
\label{tab:notation}
\begin{tabular}{ll}
\toprule
\textbf{Symbol} & \textbf{Meaning} \\
\midrule
$A$ & Binary treatment variable \\
$X$ & Pre-treatment covariates \\
$Y$ & Outcome \\
$\tau(x)$ & Individual treatment effect \\
$e(x)$ & Propensity score \\
$\mu_a(x)$ & Outcome regression function, i.e., $\mu_a(x) = \mathbb{E}[Y \mid X=x, A=a]$ for $a=0,1$ \\
$\delta(\hat{\tau}_1, \hat{\tau}_2)$ & Relative error between estimator $\hat{\tau}_1$ and $\hat{\tau}_2$ \\
$\check{\delta}(\hat{\tau}_1, \hat{\tau}_2)$ & Estimated relative error between estimator $\hat{\tau}_1$ and $\hat{\tau}_2$ \\
$\check{e}, \check{\mu}_a, \check{\gamma}, \check{\beta}_a$ & Nuisance estimators for propensity score, conditional outcomes and their coefficients \\
$\bar{e}, \bar{\mu}_a, \bar{\gamma}, \bar{\beta}_a$ & Probability limits of propensity score, conditional outcomes and their coefficients \\
$\Phi(X)$ & The shared representation of $X$, defined in Eq.~(1) \& (2) \\
\bottomrule
\end{tabular}
\end{table}

\section{Merits of Relative Error}  \label{app-A}

There are several advantages of using relative error over absolute error.

\begin{itemize}
    \item  {\bf (1) Weaker condition}. Condition \ref{cond2} is strictly weaker than Condition \ref{cond1}. Condition \ref{cond1} requires that \emph{all nuisance parameter estimators} converge to their true values at a rate faster than $n^{-1/4}$. In contrast, 
Condition \ref{cond2} imposes a weaker requirement---only that the product of the bias, $(\tilde \mu_a(x) -\mu_a(x)) (\tilde e(x) - e(x))$, converges at a rate of order $o_{\P}(n^{-1/2})$ as well as the nuisance function estimators being consistent. This allows for cases where $\tilde e(x)$ converges at a rate of $o_{\P}(n^{-1/5})$ and $\tilde \mu_a(x)$ converges at a rate of $o_{\P}(n^{-1/3})$.

\item {\bf (2) Easier to compare multiple estimators.} When comparing two estimators $\hat \tau_1(x)$ and $\hat \tau_2(x)$ in terms of absolute error,   
although both $\hat \phi(\hat \tau_1)$ and $\hat \phi(\hat \tau_2)$  are asymptotically normal, we cannot directly construct a confidence interval for $\hat \phi(\hat \tau_1)  - \hat \phi(\hat \tau_2)$ 
due to their dependency (as they use the same test data and share the same nuisance parameter estimates). In contrast, $\hat \delta(\hat \tau_1, \hat \tau_2)$ does not suffer such a problem. 

\item {\bf  (3) Double robustness}. When we replace $o_{\P}(n^{-1/2})$ in Conditions \ref{cond1} and \ref{cond2} with $o_{\P}(1)$, both $\hat \phi(\hat \tau_1)$ and $\hat \delta(\hat \tau_1, \hat \tau_2)$ are consistent (asymptotically unbiased) under their respective conditions. Thus, from Condition \ref{cond2}, $\hat \delta(\hat \tau_1, \hat \tau_2)$ exhibits the property of double robustness, meaning it is a consistent estimator if either $\tilde e(x)$ is consistent or $\tilde \mu_a(x)$ for $a = 0, 1$ are consistent. However, $\hat \phi(\hat \tau_1)$ dose not possess this property.
    
\end{itemize}

\section{Illustration of Neural Network Structure} \label{app-ill}
\begin{figure}
    \centering
    \includegraphics[width=0.8\linewidth]{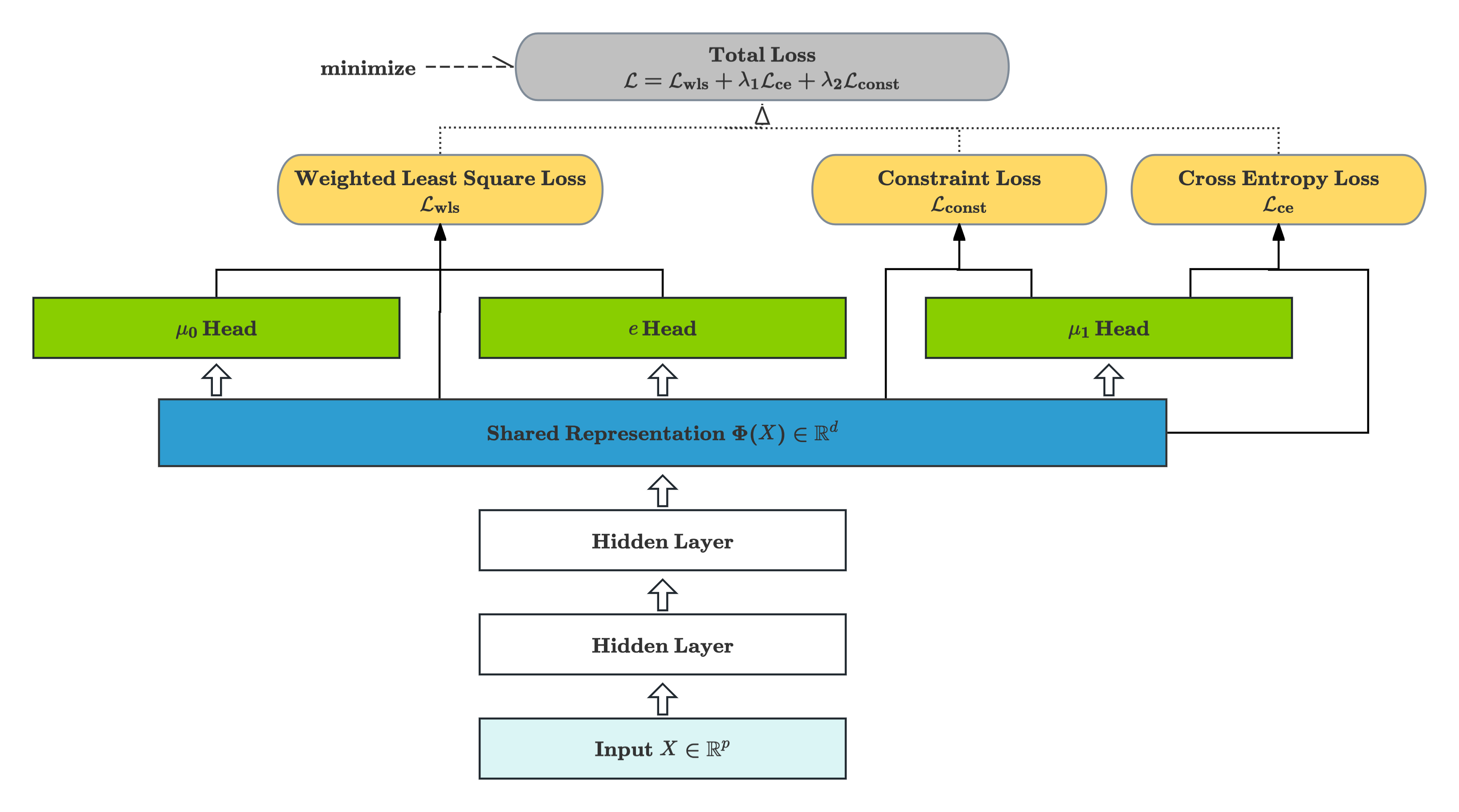}
    \caption{Neural Network Structure}
    \label{fig:network}
\end{figure}

Figure~\ref{fig:network} shows the schematic structure of our proposed network. 
The input covariates $X \in \mathbb{R}^p$ are passed through fully connected hidden layers to obtain a shared representation $\Phi(X) \in \mathbb{R}^d$. 
This representation is fed into three heads: the control outcome head $\mu_0(X)$, the treated outcome head $\mu_1(X)$, and the treatment head $e(X)$. 
The outcome heads contribute to the weighted least square loss $\mathcal{L}_{\text{wls}}$, the treatment head contributes to the cross entropy loss $\mathcal{L}_{\text{ce}}$, and the shared representation is regularized by the constraint loss $\mathcal{L}_{\text{const}}$. 
The total objective is given by
\[
\mathcal{L} = \mathcal{L}_{\mathrm{wls}} + \lambda_1 \mathcal{L}_{\mathrm{ce}} + \lambda_2 \mathcal{L}_{\mathrm{const}}.
\]
\section{Proof of Theorem~\ref{thm1}}

{\bf Theorem 1.} If the propensity score model is correctly specified, and $\check \gamma$, $\check \beta_0$ as well as $\check \beta_1$ converge to their probability limits at a rate  faster than $n^{-1/4}$, then we have  
\begin{align*}
\sqrt{n} \{\check \delta(\hat \tau_1, \hat \tau_2; \check \gamma, \check \beta_0, \check \beta_1 )  - \delta(\hat \tau_1, \hat \tau_2)\} \xrightarrow{d} \mathcal{N}(0, \sigma^{2}),   
\end{align*}
where $\sigma^{2} =  \text{Var}\{\varphi(Z; \bar u_{0}, \bar u_{1}, \bar e )\}$ and $\xrightarrow{d}$ means convergence in distribution.

\emph{Proof of Theorem 1}. As discussed in Section~\ref{sec4-1}, we first show that  
\begin{align}\label{eq:sup-c1}
\check \delta(\hat \tau_1, \hat \tau_2; \check \gamma, \check \beta_0, \check \beta_1 ) -  \check \delta(\hat \tau_1, \hat \tau_2; \bar \gamma, \bar \beta_0, \bar \beta_1 ) = o_{\P}(n^{-1/2}).    
\end{align}
By a Taylor expansion of $\check \delta(\hat \tau_1, \hat \tau_2; \check \gamma, \check \beta_0, \check \beta_1 )$ around $(\bar \gamma, \bar \beta_0, \bar \beta_1 )$, we obtain
\begin{align*}
    &\check \delta(\hat \tau_1, \hat \tau_2; \check \gamma, \check \beta_0, \check \beta_1 )   - \check \delta(\hat \tau_1, \hat \tau_2; \bar \gamma, \bar \beta_0, \bar \beta_1 )  \\
    = & \Delta_\gamma^\intercal   ( \check \gamma - \bar \gamma) +  \Delta_{\beta_0}^\intercal (\check \beta_0 - \bar \beta_0)  + \Delta_{\beta_1}^\intercal (\check \beta_1 - \bar \beta_1) + O_{\P}( (\check \gamma - \bar \gamma)^2 +  (\check \gamma - \bar \gamma) ( \check \beta_1 - \bar \beta_1) +  (\check \gamma - \bar \gamma) ( \check \beta_0 - \bar \beta_0)   ),  
\end{align*} 
where 
\begin{align*}
   \Delta_\gamma  ={}& - \frac 1 n \sum_{i=1}^n  2 ( \hat \tau_1(X_i) - \hat \tau_2(X_i)  )  \Big ( \frac{ A_i (1 - \bar e(X_i)) (Y_i - \bar \mu_1(X_i)) }{ \bar e(X_i) } +   \frac{ (1-A_i) \bar e(X_i)  (Y_i - \bar \mu_0(X_i))  }{ 1 - \bar e(X_i) } \Big )  \Phi_1(X_i), \\
    \Delta_{\beta_0} ={}&  -   \frac 1 n \sum_{i=1}^n   2 ( \hat \tau_1(X_i)  -\hat \tau_2(X_i)  ) \left ( 1  - \frac{1- A_i  }{1- \bar e(X_i) } \right )  \Phi_2(X_i), \\
       \Delta_{\beta_1} ={}&   \frac 1 n \sum_{i=1}^n   2 ( \hat \tau_1(X_i)  -\hat \tau_2(X_i)  )  \left ( 1 -  \frac{ A_i  }{ \bar e(X_i) }  \right )  \Phi_2(X_i). 
\end{align*}
By the condition that $\check{\gamma} = \bar \gamma + o_{\mathrm{\P}}(n^{-1/4})$, $\check{\beta}_0 = \bar \beta_0 + o_{\mathrm{\P}}(n^{-1/4})$ and $\check{\beta}_0 = \bar \beta_0 + o_{\mathrm{\P}}(n^{-1/4})$, we obtain
$
O_{\P}( (\check \gamma - \bar \gamma)^2 +  (\check \gamma - \bar \gamma) ( \check \beta_1 - \bar \beta_1) +  (\check \gamma - \bar \gamma) ( \check \beta_0 - \bar \beta_0) ) = o_{\P}(n^{-1/2})
$. Thus, we only need to deal with $\Delta_\gamma^\intercal   ( \check \gamma - \bar \gamma)$, $\Delta_{\beta_0}^\intercal (\check \beta_0 - \bar \beta_0)  $ and $\Delta_{\beta_1}^\intercal (\check \beta_1 - \bar \beta_1)$. 

Since $\bar \gamma$, $\bar \beta_0$ and $\bar \beta_1$ are the probability limits of $\check \gamma$, $\check \beta_0$ and $\check \beta_1$, respectively, we obtain $\check \gamma - \bar \gamma = o_{\P}(1)$, $\check \beta_0 - \bar \beta_0 = o_{\P}(1)$ and $\check \beta_1 - \bar \beta_1 = o_{\P}(1)$. 

Then, it sufficies to show that $\Delta_\gamma = O_{\P}(n^{-1/2})$, $\Delta_{\beta_0} = O_{\P}(n^{-1/2})$ and $\Delta_{\beta_1} = O_{\P}(n^{-1/2})$. By CLT,  $\Delta_\gamma - \E(\Delta_\gamma) = O_{\P}(n^{-1/2})$, $\Delta_{\beta_{0}} - \E(\Delta_{\beta_{0}}) = O_{\P}(n^{-1/2})$ and $\Delta_{\beta_{1}} - \E(\Delta_{\beta_{1}}) = O_{\P}(n^{-1/2})$, then, we only need to show that $ \E(\Delta_\gamma) = \E(\Delta_{\beta_{0}}) = \E(\Delta_{\beta_{1}}) = 0$.

We first deal with $\E(\Delta_\gamma)$.
\begin{align*}
\E(\Delta_\gamma) = &\E \left( - \frac 1 n \sum_{i=1}^n  2 ( \hat \tau_1(X_i) - \hat \tau_2(X_i)  )  \Big ( \frac{ A_i (1 - \bar e(X_i)) (Y_i - \bar \mu_1(X_i)) }{ \bar e(X_i) } +   \frac{ (1-A_i) \bar e(X_i)  (Y_i - \bar \mu_0(X_i))  }{ 1 - \bar e(X_i) } \Big )  \Phi_1(X_i) \right) \\
= & 2\E \left( ( \hat \tau_1(X) - \hat \tau_2(X)  )  \Big ( \frac{ A (1 - \bar e(X)) (Y - \bar \mu_1(X)) }{ \bar e(X) } +   \frac{ (1-A) \bar e(X)  (Y - \bar \mu_0(X))  }{ 1 - \bar e(X) } \Big )  \Phi_1(X) \right)\\
= & 2\E \left( ( \hat \tau_1(X) - \hat \tau_2(X)  )   \frac{ A (1 - \bar e(X)) (Y - \bar \mu_1(X)) }{ \bar e(X) }  \Phi_1(X) \right)\\
& + 2\E \left(  ( \hat \tau_1(X) - \hat \tau_2(X)  ) \frac{ (1-A) \bar e(X)  (Y - \bar \mu_0(X))  }{ 1 - \bar e(X) } \Phi_1(X) \right)\\
=& 0.
\end{align*}
The last equation holds by the definition of $\bar \beta_{0}$ and $\bar \beta_{1}$ and the fact that $\Phi_1(X)$ is a sub-vector of $\Phi_2(X)$. 

We then deal with $\E(\Delta_{\beta_{0}})$.
\begin{align*}
\E(\Delta_{\beta_0}) ={}& \E \left( -   \frac 1 n \sum_{i=1}^n   2 ( \hat \tau_1(X_i)  -\hat \tau_2(X_i)  ) \left ( 1  - \frac{1- A_i  }{1- \bar e(X_i) } \right )  \Phi_2(X_i) \right) \\
= & - 2\E \left(  ( \hat \tau_1(X)  -\hat \tau_2(X)  ) \left ( 1  - \frac{1- A  }{1- \bar e(X) } \right )  \Phi_2(X) \right)\\
= & - 2 \E_{X}\left( \left.\E \left(  ( \hat \tau_1(X)  -\hat \tau_2(X)  ) \left ( 1  - \frac{1- A  }{1- \bar e(X) } \right )  \Phi_2(X) \right \vert X \right) \right )\\
= & 0.
\end{align*}
The last equation holds since the PS model is correct. Finally, we handle $\E(\Delta_{\beta_{1}})$.
\begin{align*}
\E(\Delta_{\beta_1}) ={}&  \E \left( \frac 1 n \sum_{i=1}^n   2 ( \hat \tau_1(X_i)  -\hat \tau_2(X_i)  )  \left ( 1 -  \frac{ A_i  }{ \bar e(X_i) }  \right )  \Phi_2(X_i) \right )\\
= & 2  \E \left( ( \hat \tau_1(X)  -\hat \tau_2(X)  )  \left ( 1 -  \frac{ A  }{ \bar e(X) }  \right )  \Phi_2(X) \right )\\
= & 2 \E_{X} \left ( \E \left( \left. ( \hat \tau_1(X)  -\hat \tau_2(X)  )  \left ( 1 -  \frac{ A  }{ \bar e(X) }  \right )  \Phi_2(X) \right \vert X \right ) \right)\\
=&0.
\end{align*}
The last equation holds since the PS model is correct. Therefore, equation~\eqref{eq:sup-c1} holds.

We then want to show that 
\begin{align}\label{eq:norm}
  \sqrt{n} ( \check \delta(\hat \tau_1, \hat \tau_2; \bar \gamma, \bar \beta_0, \bar \beta_1 ) -  \delta(\hat \tau_1, \hat \tau_2) ) \rightarrow \mathcal N (0, \sigma^2).
\end{align}
By definiton,
\begin{align*}
    &  \check \delta(\hat \tau_1, \hat \tau_2; \bar \gamma, \bar \beta_0, \bar \beta_1 )  \\
= &  {} \frac {1}{n} \sum_{i=1}^n \{\hat \tau_1^2(X_i)  - \hat \tau_2^2(X_i)     \} \\
    -{}& \frac {1}{n} \sum_{i=1}^n \left (  2 \{ \hat \tau_1(X_i)  -\hat \tau_2(X_i)  \} \cdot \left [ \frac{ A_i \{Y_i - \bar \mu_1(X_i)\}  }{ \bar e(X_i) } + \bar \mu_1(X_i)  - \frac{ (1-A_i) \{Y_i - \bar \mu_0(X_i)\} }{ 1 - \bar e(X_i) } -  \bar \mu_0(X_i)    \right ]    \right )\\
\end{align*}
If model~\eqref{eq1} is correct, 
\begin{align*}
    & \E \left [ \{\hat \tau_1^2(X_i)   - \hat \tau_2^2(X_i) \} \right ] \\
    -{}&\E \left \{ \left (  2 \{ \hat \tau_1(X_i)  -\hat \tau_2(X_i)  \} \cdot \left [ \frac{ A_i \{Y_i - \bar \mu_1(X_i)\}  }{ \bar e(X_i) } + \bar \mu_1(X_i)  - \frac{ (1-A_i) \{Y_i - \bar \mu_0(X_i)\} }{ 1 - \bar e(X_i) } -  \bar \mu_0(X_i)    \right ]    \right ) \right \}\\
= &  {} \E \left [ \{\hat \tau_1^2(X)   - \hat \tau_2^2(X) \} \right ] \\
    -{}& \left.\E_{X_{i}} \E \left [\left \{ \left (  2 \{ \hat \tau_1(X_i)  -\hat \tau_2(X_i)  \} \cdot \left [ \frac{ A_i \{Y_i - \bar \mu_1(X_i)\}  }{ \bar e(X_i) } + \bar \mu_1(X_i)  - \frac{ (1-A_i) \{Y_i - \bar \mu_0(X_i)\} }{ 1 - \bar e(X_i) } -  \bar \mu_0(X_i)    \right ]    \right ) \right \} \right \vert X_{i} \right]\\
= &  {} \E \left [ \{\hat \tau_1^2(X)   - \hat \tau_2^2(X) \} \right ] \\
    -{}&  \E_{X_{i}} \E \left [\left \{ \left (  2 \{ \hat \tau_1(X_i)  -\hat \tau_2(X_i)  \} \cdot \left [ \frac{ A_i \{Y_i(1) - \bar \mu_1(X_i)\}  }{ \bar e(X_i) } + \bar \mu_1(X_i) \right. \right. \right. \right. \\
    & \left.\left.\left.\left.\left. - \frac{ (1-A_i) \{Y_i(0) - \bar \mu_0(X_i)\} }{ 1 - \bar e(X_i) } -  \bar \mu_0(X_i)    \right ]    \right ) \right \} \right \vert X_{i} \right]\\
= &  {} \E \left [ \{\hat \tau_1^2(X)   - \hat \tau_2^2(X) \} \right ] \\
    -{}& \E_{X_{i}} \E \left [\left \{ \left (  2 \left \{ \hat \tau_1(X_i)  -\hat \tau_2(X_i)  \} \cdot \left [ \{Y_i(1) - \bar \mu_1(X_i)\} + \bar \mu_1(X_i)  - \{Y_i(0) - \bar \mu_0(X_i)\}  -  \bar \mu_0(X_i)    \right ]    \right ) \right \} \right \vert X_{i} \right]\\
= &  {} \E \left [ \{\hat \tau_1^2(X)   - \hat \tau_2^2(X) \} \right ] - \E_{X_{i}} \E  [ \{  (  2 \{ \hat \tau_1(X_i)  -\hat \tau_2(X_i)  \}\tau(X_{i})|X_{i}]\\
= &  {} \E \left [ \{\hat \tau_1^2(X)   - \hat \tau_2^2(X) \} -  2 \{ \hat \tau_1(X)  -\hat \tau_2(X)  \}\tau(X)\right ]\\
= & \delta(\hat \tau_1, \hat \tau_2).
\end{align*}

If model~\eqref{eq2} is correct,
\begin{align*}
    & \E \left [  \{\hat \tau_1^2(X_i)   - \hat \tau_2^2(X_i) \} \right ] \\
    -{}&\E \left \{  \left (  2 \{ \hat \tau_1(X_i)  -\hat \tau_2(X_i)  \} \cdot \left [ \frac{ A_i \{Y_i - \bar \mu_1(X_i)\}  }{ \bar e(X_i) } + \bar \mu_1(X_i)  - \frac{ (1-A_i) \{Y_i - \bar \mu_0(X_i)\} }{ 1 - \bar e(X_i) } -  \bar \mu_0(X_i)    \right ]    \right ) \right \}\\
= &  {} \E \left [ \{\hat \tau_1^2(X)   - \hat \tau_2^2(X) \} \right ] \\
    -{}& \left. \E_{X_{i}} \E \left [\left \{ \left (  2 \{ \hat \tau_1(X_i)  -\hat \tau_2(X_i)  \} \cdot \left [ \frac{ A_i \{Y_i - \bar \mu_1(X_i)\}  }{ \bar e(X_i) } + \bar \mu_1(X_i)  - \frac{ (1-A_i) \{Y_i - \bar \mu_0(X_i)\} }{ 1 - \bar e(X_i) } -  \bar \mu_0(X_i)    \right ]    \right ) \right \} \right \vert X_{i} \right]\\
= &  {} \E \left [ \{\hat \tau_1^2(X)   - \hat \tau_2^2(X) \} \right ] \\
    -{}&  \E_{X_{i}} \E \left [\left \{ \left (  2 \{ \hat \tau_1(X_i)  -\hat \tau_2(X_i)  \} \cdot \left [ \frac{ A_i \{Y_i(1) - \bar \mu_1(X_i)\}  }{ \bar e(X_i) } + \bar \mu_1(X_i) \right. \right. \right. \right. \\
    & \left.\left.\left.\left.\left. - \frac{ (1-A_i) \{Y_i(0) - \bar \mu_0(X_i)\} }{ 1 - \bar e(X_i) } -  \bar \mu_0(X_i)    \right ]    \right ) \right \} \right \vert X_{i} \right]\\
= &  {} \E \left [ \{\hat \tau_1^2(X)   - \hat \tau_2^2(X) \} \right ] \\
    -{}& \E_{X_{i}} \E \left [\left \{ \left (  2 \{ \hat \tau_1(X_i)  -\hat \tau_2(X_i)  \} \cdot \left [ \frac{ A_i \{\bar \mu_1(X_i) - \bar \mu_1(X_i)\}  }{ \bar e(X_i) } + \bar \mu_1(X_i) \right. \right. \right. \right. \\
    & \left.\left.\left.\left.\left. - \frac{ (1-A_i) \{\bar \mu_0(X_i) - \bar \mu_0(X_i)\} }{ 1 - \bar e(X_i) } -  \bar \mu_0(X_i)    \right ]    \right ) \right \} \right \vert X_{i} \right]\\
= &  {} \E \left [ \{\hat \tau_1^2(X)   - \hat \tau_2^2(X) \} \right ] - \E_{X_{i}} \E  [ \{  (  2 \{ \hat \tau_1(X_i)  -\hat \tau_2(X_i)  \}\tau(X_{i})|X_{i}]\\
= &  {} \E \left [ \{\hat \tau_1^2(X)   - \hat \tau_2^2(X) \} -  2 \{ \hat \tau_1(X)  -\hat \tau_2(X)  \}\tau(X)\right ]\\
= & \delta(\hat \tau_1, \hat \tau_2).
\end{align*}
Therefore, when at least one of models~\eqref{eq1} or ~\eqref{eq2} is correct, $\bar \delta(\hat \tau_1, \hat \tau_2; \bar \gamma, \bar \beta_0, \bar \beta_1 ) -  \delta(\hat \tau_1, \hat \tau_2)$ is the average of i.i.d. observations with mean 0 and variance $ \text{Var}\{\varphi(Z; \bar u_{0}, \bar u_{1}, \bar e )\}$. By CLT,~\ref{eq:norm} holds with $\sigma^{2} =  \text{Var}\{\varphi(Z; \bar u_{0}, \bar u_{1}, \bar e )\}$.

\hfill $\Box$

\subsection{Proof of Proposition~\ref{prop}}

{\bf Proposition 2.} 
Under the conditions in Theorem~\ref{thm1},
a consistent estimator of $\sigma^{2}$ is 
\begin{align*}
\hat{\sigma}^{2} = \frac{1}{n} \sum^{n}_{i=1}\left\{\varphi(Z_{i}; \check u_{0}, \check u_{1}, \check e ) -  \check \delta(\hat \tau_1, \hat \tau_2; \check \gamma, \check \beta_0, \check \beta_1 ) \right\}^{2}, 
\end{align*}
an asymptotic $(1-\eta)$ confidence interval for $\delta(\hat \tau_1, \hat \tau_2)$ is $\check \delta(\hat \tau_1, \hat \tau_2; \check \gamma, \check \beta_0, \check \beta_1 ) \pm z_{\eta/2}\sqrt{\hat{\sigma}^2/n}$, where $z_{\eta/2}$ is the $(1-\eta/2)$ quantile of the standard normal distribution.

\begin{proof}[Proof of Proposition 2] 
\begin{align*}
 \hat{\sigma}^{2} - \sigma^2
= & \frac{1}{n} \sum^{n}_{i=1}\left\{\varphi(Z_{i}; \check u_{0}, \check u_{1}, \check e )  -  \check \delta(\hat \tau_1, \hat \tau_2; \check \gamma, \check \beta_0, \check \beta_1 ) \right\}^{2} - \E \left\{\varphi(Z_{i}; \bar u_{0}, \bar u_{1}, \bar e )  -  \E \varphi(Z_{i}; \bar u_{0}, \bar u_{1}, \bar e ) \right\}^{2}\\
= & \underbrace{\frac{1}{n} \sum^{n}_{i=1}\left\{\varphi(Z_{i}; \check u_{0}, \check u_{1}, \check e )  -  \check \delta(\hat \tau_1, \hat \tau_2; \check \gamma, \check \beta_0, \check \beta_1 ) \right\}^{2} - \frac{1}{n} \sum^{n}_{i=1} \left\{\varphi(Z_{i}; \bar u_{0}, \bar u_{1}, \bar e )  -  \E \varphi(Z_{i}; \bar u_{0}, \bar u_{1}, \bar e ) \right\}^{2}}_{\textcircled{1}}\\
& \underbrace{+ \frac{1}{n} \sum^{n}_{i=1} \left\{\varphi(Z_{i}; \bar u_{0}, \bar u_{1}, \bar e )  -  \E \varphi(Z_{i}; \bar u_{0}, \bar u_{1}, \bar e ) \right\}^{2} -\E \left\{\varphi(Z_{i}; \bar u_{0}, \bar u_{1}, \bar e )  -  \E \varphi(Z_{i}; \bar u_{0}, \bar u_{1}, \bar e ) \right\}^{2}}_{\textcircled{2}}.
\end{align*}
By the law of large numbers (LLN), $\textcircled{2}\overset{p}{\rightarrow} 0$, we only need to deal with $\textcircled{1}$:
\begin{align*}
&\frac{1}{n} \sum^{n}_{i=1}\left\{\varphi(Z_{i}; \check u_{0}, \check u_{1}, \check e )  -  \check \delta(\hat \tau_1, \hat \tau_2; \check \gamma, \check \beta_0, \check \beta_1 ) \right\}^{2}\\ 
= & \frac{1}{n} \sum^{n}_{i=1}\left\{\varphi(Z_{i}; \check u_{0}, \check u_{1}, \check e ) - \varphi(Z_{i}; \bar u_{0}, \bar u_{1}, \bar e ) + \varphi(Z_{i}; \bar u_{0}, \bar u_{1}, \bar e ) -  \check \delta(\hat \tau_1, \hat \tau_2; \check \gamma, \check \beta_0, \check \beta_1 ) \right\}^{2}\\ 
= & \underbrace{\frac{1}{n} \sum^{n}_{i=1}\left\{\varphi(Z_{i}; \check u_{0}, \check u_{1}, \check e ) - \varphi(Z_{i}; \bar u_{0}, \bar u_{1}, \bar e ) \right \}^{2}}_{\textcircled{a}} + \underbrace{\frac{1}{n} \sum^{n}_{i=1} \left \{ \varphi(Z_{i}; \bar u_{0}, \bar u_{1}, \bar e ) -  \check \delta(\hat \tau_1, \hat \tau_2; \check \gamma, \check \beta_0, \check \beta_1 ) \right\}^{2}}_{\textcircled{b}}\\
& + \underbrace{\frac{2}{n} \sum^{n}_{i=1} \left\{\varphi(Z_{i}; \check u_{0}, \check u_{1}, \check e ) - \varphi(Z_{i}; \bar u_{0}, \bar u_{1}, \bar e ) \right \}  \left \{ \varphi(Z_{i}; \bar u_{0}, \bar u_{1}, \bar e ) -  \check \delta(\hat \tau_1, \hat \tau_2; \check \gamma, \check \beta_0, \check \beta_1 ) \right\}}_{\textcircled{c}}.
\end{align*}
By LLN, $\textcircled{a}\overset{p}{\rightarrow} 0$, and $\textcircled{c}\overset{p}{\rightarrow} 0$. Similarly, we can obtain 
\begin{align*}
\textcircled{b} = \frac{1}{n} \sum^{n}_{i=1} \left\{\varphi(Z_{i}; \bar u_{0}, \bar u_{1}, \bar e )  -  \E \varphi(Z_{i}; \bar u_{0}, \bar u_{1}, \bar e ) \right\}^{2} + o_{p}(1).    
\end{align*}
Therefore $\textcircled{1} \overset{p}{\rightarrow} 0$, which leads to $ \hat{\sigma}^{2} \overset{p}{\rightarrow} \sigma^2$. 

The asymptotic $1-\eta$ confidence interval is constructed by the standard theory.        

\end{proof}

\section{Experimental Details}
\label{appendix:experiments}
\subsection{Dataset Details} 
\label{app-B1}
\textbf{IHDP.} The IHDP dataset is based on a randomized controlled trial conducted as part of the Infant Health and Development Program. The goal is to assess the impact of specialist home visits on children’s future cognitive outcomes. Following Hill~\citep{Hill01012011}, a subset of treated units is removed to introduce selection bias, creating a semi-synthetic evaluation setting. The dataset contains 747 samples (139 treated and 608 control), each with 25 pre-treatment covariates. The simulated outcome is  the same as that in Shalit et al.(2017)~\citep{pmlr-v70-shalit17a}, by setting “A” in the NPCI package~\citep{dorie2016npci}.

\textbf{Twins.} The Twins dataset is constructed from twin births in the U.S.. For each twin pair, the heavier twin is assigned as the treated unit ($t_i = 1$), and the lighter twin as the control ($t_i = 0$). We extract 28 covariates related to parental, pregnancy, and birth characteristics from the original data and generate an additional 10 covariates following \citep{wu2022learning}. The outcome of interest is the one-year mortality of each child. We restrict the analysis to same-sex twins with birth weights below 2000g and without any missing features, yielding a final dataset with 5,271 samples. 
The treatment assignment mechanism is defined as:
\(
t_i \mid x_i \sim \mathrm{Bern}\left( \sigma(w^\top X + n) \right),
\)
where $\sigma(\cdot)$ is the sigmoid function, $w^\top \sim \mathcal{U}((-0.1, 0.1)^{38 \times 1})$, and $n \sim \mathcal{N}(0, 0.1)$.

\textbf{Jobs.} The Jobs dataset is a standard benchmark in causal inference, originally introduced by LaLonde (1986)~\citep{9c8bcb59-753a-33cb-871e-17f05b11d793}. It evaluates the impact of job training on employment outcomes by combining data from a randomized study (National Supported Work program) with observational records (PSID), following the setup of Smith and Todd (2005)~\citep{ASMITH2005305}. The dataset includes 297 treated units, 425 control units from the experimental sample, and 2490 control units from the observational sample. Each record consists of 8 covariates, such as age, education, ethnicity, and pre-treatment earnings. The task is framed as a binary classification problem predicting unemployment status post-treatment, using features defined by Dehejia and Wahba (2002)~\citep{10.1162/003465302317331982}.
\subsection{Choosing Different Nuisance Estimators}
\label{app-gao}
\textbf{Experimental Set-up.} As for the nuisance estimators, we choose linear regression and gradient boosting as Gao~\citep{Gao2024} used in their paper. The evaluation metrics are the same as those in Section~\ref{sec5}. We provide the results on the IHDP and the Twins.

\textbf{Experimental Results.} Table~\ref{tab:comparison} summarizes the results on the IHDP and Twins datasets. When plugging conventional nuisance estimators (linear regression and gradient boosting) into the relative error framework, the resulting procedures do achieve nominal coverage. Nevertheless, the corresponding variance is so large that the confidence intervals frequently include zero, making it essentially impossible to tell which candidate estimator is superior. These baselines therefore serve as valid but uninformative references. In contrast, our proposed method not only maintains well-calibrated coverage but also delivers much higher selection accuracy, producing confidence intervals that are substantially tighter and practically useful for identifying the winner.
\begin{table}[t]
\centering
\caption{Relative Error Estimation Performance with Different Nuisance Estimators on the IHDP and Twins datasets.}
\label{tab:comparison}
\resizebox{\textwidth}{!}{
\begin{tabular}{l|cc|cc}
\toprule
\multicolumn{1}{l|}{} &
\multicolumn{2}{c|}{\textbf{IHDP}} & \multicolumn{2}{c}{\textbf{Twins}} \\
\cmidrule(lr){2-3} \cmidrule(lr){4-5}
\textbf{Nuisance Estimators} & Coverage Rate & Selection Accuracy & Coverage Rate & Selection Accuracy \\
\midrule
Linear Regression & 0.94 & 0.44 & 0.94 & 0.88 \\
Gradient Boosting  & 0.95 & 0.48 & 0.94 & 0.86 \\
\midrule
\textbf{Ours} & \textbf{0.96} & \textbf{0.80} & \textbf{0.94} & \textbf{0.94} \\
\bottomrule
\end{tabular}
}
\vspace{-8pt}
\end{table}
\subsection{Results on Jobs}
\label{app-B2}
\textbf{Evaluation Metrics.}
For the \textbf{Jobs} datasets, as there are no counterfactual outcomes, we report the true Average Treatment Effect on the Treated (ATT) and the Policy Risk (\(\mathcal{R}_{\text{pol}}\)) recommended by Shalit et al.\citep{shalit2017estimatingindividualtreatmenteffect}.
Specifically, the policy risk can be estimated using only the randomized subset of the Jobs dataset:
\[
\hat{\mathcal{R}}{\text{pol}} = 1 - \left( \frac{1}{|A_1 \cap T_1 \cap E|} \sum_{x_i \in A_1 \cap T_1 \cap E} y_1^{(i)} \cdot \frac{|A_1 \cap E|}{|E|} + \frac{1}{|A_0 \cap T_0 \cap E|} \sum_{x_i \in A_0 \cap T_0 \cap E} y_0^{(i)} \cdot \frac{|A_0 \cap E|}{|E|} \right)
\]
where E denotes units from the experimental group, \(A_1 = \{x_i : \hat{y}_1^{(i)} - \hat{y}_0^{(i)} > 0\}, A_0 = \{x_i : \hat{y}_1^{(i)} - \hat{y}_0^{(i)} < 0\}\), and \(T_1, T_0\) are the treated and control subsets, respectively. Since all treated units $T$ belong to the randomized subset $E$, the true Average Treatment Effect on the Treated (ATT) can be identified and computed as:
\[
\text{ATT} = \frac{1}{|T|} \sum_{i \in T} y_i - \frac{1}{|C \cap E|} \sum_{i \in C \cap E} y_i
\]
where C denotes the control group. We evaluate estimation accuracy using the ATT error:
\(\epsilon_{\text{ATT}} = \left| \text{ATT} - \frac{1}{|T|} \sum_{i \in T} \left( f(x_i, 1) - f(x_i, 0) \right) \right|.\)

\textbf{Accuracy of the CATE Estimation.} We evaluate the performance of CATE estimation by our network and compare it with baselines mentioned in Section~\ref{sec5}. We average over 20 realizations of our network, and the results are presented in Table~\ref{tab:jobs}. One can clearly see that our proposed method achieves the best performance across all metrics, having the lowest \(\hat{\mathcal{R}}{\text{pol}}\) and \(\epsilon_{\text{ATT}}\) in both training sets and test sets.
\begin{table}[t]
\centering
\caption{Performance on the Jobs dataset (in-sample and out-of-sample).}
\label{tab:jobs}
\resizebox{0.75\textwidth}{!}{
\begin{tabular}{l|cc|cc}
\toprule
\multirow{2}{*}{\textbf{Method}} 
& $\mathcal{R}_{\text{pol}}^{\text{in}}$ & $\epsilon_{\text{ATT}}^{\text{in}}$ 
& $\mathcal{R}_{\text{pol}}^{\text{out}}$ & $\epsilon_{\text{ATT}}^{\text{out}}$ \\
\midrule
LinDML     & 0.158 $\pm$ 0.015 & 0.019 $\pm$ 0.015 & 0.183 $\pm$ 0.040 & \textbf{0.053 $\pm$ 0.051} \\
SpaDML     & 0.150 $\pm$ 0.024 & 0.131 $\pm$ 0.118 & 0.165 $\pm$ 0.046 & 0.144 $\pm$ 0.134 \\
CForest    & 0.114 $\pm$ 0.016 & 0.025 $\pm$ 0.018 & 0.155 $\pm$ 0.028 & 0.058 $\pm$ 0.047 \\
X-Learner  & 0.169 $\pm$ 0.037 & 0.026 $\pm$ 0.015 & 0.173 $\pm$ 0.034 & \textbf{0.053 $\pm$ 0.050} \\
S-Learner  & 0.148 $\pm$ 0.026 & 0.095 $\pm$ 0.040 & 0.160 $\pm$ 0.027 & 0.115 $\pm$ 0.070 \\
TarNet     & 0.141 $\pm$ 0.005 & 0.183 $\pm$ 0.047 & 0.145 $\pm$ 0.009 & 0.190 $\pm$ 0.074 \\
Dragonnet  & 0.230 $\pm$ 0.011 & 0.021 $\pm$ 0.018 & 0.143 $\pm$ 0.009 & 0.172 $\pm$ 0.039\\
DRCFR      & 0.142 $\pm$ 0.005 & 0.122 $\pm$ 0.017 & 0.218 $\pm$ 0.021 & 0.048 $\pm$ 0.032 \\
SCIGAN     & 0.144 $\pm$ 0.005 & 0.112 $\pm$ 0.025 & 0.220 $\pm$ 0.026 & 0.049 $\pm$ 0.034 \\
DESCN      & 0.192 $\pm$ 0.029 & 0.098 $\pm$ 0.029 & 0.143 $\pm$ 0.011 & 0.065 $\pm$ 0.046 \\
ESCFR      & 0.202 $\pm$ 0.023 & 0.086 $\pm$ 0.028 & 0.145 $\pm$ 0.011 & 0.076 $\pm$ 0.045 \\
\midrule
\textbf{Ours} & \textbf{0.112 $\pm$ 0.019} & \textbf{0.018 $\pm$ 0.012} & \textbf{0.131 $\pm$ 0.030} & \textbf{0.053 $\pm$ 0.039} \\
\bottomrule
\end{tabular}
}
\end{table}
\textbf{Sensitive Analysis and Ablation Study.} We explore which value of \(\lambda_1\), \(\lambda_2\) and \(\rho\) can achieve the best performance. The results are demonstrated in Table~\ref{tab:sensitive_jobs_combined}. We can see that our model is not sensitive to the change of hyperparameters. That is, the performance of CATE estimation remains relatively stable across a range of hyperparameters. For the ablation study presented in Table~\ref{tab:ablation studies_3}, as that in IHDP and Twins, taking \(\mathcal{L}_{\text{ce}}\) off only causes a moderate decline, while removing \(\mathcal{L}_{\text{cosnt}}\) brings a whole disaster.

\begin{table}[t]
\centering
\caption{Sensitivity analysis on the Jobs dataset with respect to $\lambda_1$, $\lambda_2$, and $\rho$.}
\label{tab:sensitive_jobs_combined}
\resizebox{\textwidth}{!}{
\begin{tabular}{c|cccc|c|cccc|c|cccc}
\toprule
\multicolumn{5}{c|}{\textbf{$\lambda_1$}} & \multicolumn{5}{c|}{\textbf{$\lambda_2$}} & \multicolumn{5}{c}{\textbf{$\rho$}} \\
\cmidrule(lr){1-5} \cmidrule(lr){6-10} \cmidrule(lr){11-15}
\textbf{Value} & $\mathcal{R}_{\text{pol}}^{\text{in}}$ & $\epsilon_{\text{ATT}}^{\text{in}}$ & $\mathcal{R}_{\text{pol}}^{\text{out}}$ & $\epsilon_{\text{ATT}}^{\text{out}}$ 
& \textbf{Value} & $\mathcal{R}_{\text{pol}}^{\text{in}}$ & $\epsilon_{\text{ATT}}^{\text{in}}$ & $\mathcal{R}_{\text{pol}}^{\text{out}}$ & $\epsilon_{\text{ATT}}^{\text{out}}$ 
& \textbf{Value} & $\mathcal{R}_{\text{pol}}^{\text{in}}$ & $\epsilon_{\text{ATT}}^{\text{in}}$ & $\mathcal{R}_{\text{pol}}^{\text{out}}$ & $\epsilon_{\text{ATT}}^{\text{out}}$ \\
\midrule
0.1 & 0.113 & 0.018 & 0.132 & 0.054
    & 0.1 & 0.109 & 0.021 & 0.129 & 0.056
    & 10 & 0.124 & 0.020 & 0.141 & 0.051 \\
0.5 & 0.113 & 0.022 & 0.130 & 0.058
    & 0.5 & 0.109 & 0.020 & 0.128 & 0.054
    & 50 & 0.112 & 0.019 & 0.131 & 0.052 \\
\textbf{1} & \textbf{0.112} & \textbf{0.018} & \textbf{0.131} & \textbf{0.053}
    & \textbf{1} & \textbf{0.112} & \textbf{0.018} & \textbf{0.131} & \textbf{0.053}
    & \textbf{100} & \textbf{0.112} & \textbf{0.018} & \textbf{0.131} & \textbf{0.053} \\
2 & 0.115 & 0.020 & 0.135 & 0.054
  & 2 & 0.117 & 0.019 & 0.135 & 0.050
  & 200 & 0.115 & 0.020 & 0.132 & 0.053 \\
10 & 0.121 & 0.027 & 0.140 & 0.060
   & 10 & 0.123 & 0.027 & 0.144 & 0.060
   & 1000 & 0.114 & 0.020 & 0.133 & 0.052 \\
\bottomrule
\end{tabular}
}
\end{table}

\begin{table}[H]
    \centering
    \caption{ablation studies Jobs}
    \label{tab:ablation studies_3}
    \begin{tabular}{c|cccccc}
    \toprule
        Training Loss & $\mathcal{R}_{\text{pol}}^{\text{within-s.}}$ & $\epsilon_{\text{ATT}}^{\text{within-s.}}$ & $\mathcal{R}_{\text{pol}}^{\text{out-of-s.}}$ & $\epsilon_{\text{ATT}}^{\text{out-of-s.}}$ \\
    \midrule
    $\mathcal{L}_{\text{wls}}$ \& $\mathcal{L}_{\text{const}}$ & 0.114 & 0.023  & 0.134 & 0.053  \\
    $\mathcal{L}_{\text{wls}}$ \& $\mathcal{L}_{\text{ce}}$ & 0.121 & 0.029 & 0.141 & 0.055  \\
    \bottomrule
    \textbf{Full (Ours)} & \textbf{0.112} & \textbf{0.018} & \textbf{0.131} & \textbf{0.053} \\
    \bottomrule
    \end{tabular}

\end{table}

\subsection{Extended Sensitive Analysis}
\label{app-B3}
In this section we present the results of the sensitive analysis of hyperparameter \(\lambda_1\) and \(\rho\) in the IHDP and Twins dataset. One can see from Table~\ref{tab:sensitivity_lambda1} and Table~\ref{tab:sensitivity_rho} that our model is robust to the change of \(\lambda_1\) and \(\rho\), remaining good performance in the CATE estimation as well as relative error prediction.
\begin{table}[t]
\centering
\caption{Sensitivity analysis of $\lambda_1$ on IHDP and Twins datasets.}
\label{tab:sensitivity_lambda1}
\resizebox{\textwidth}{!}{
\begin{tabular}{c|cccccc|c|cccccc}
\toprule
\multicolumn{7}{c|}{\textbf{IHDP}} & \multicolumn{7}{c}{\textbf{Twins}} \\
\cmidrule(lr){1-7} \cmidrule(lr){8-14}
\textbf{Value} 
& $\sqrt{\epsilon_{\text{PEHE}}^{\text{in}}}$ & $\epsilon_{\text{ATE}}^{\text{in}}$ 
& $\sqrt{\epsilon_{\text{PEHE}}^{\text{out}}}$ & $\epsilon_{\text{ATE}}^{\text{out}}$ 
& Coverage & Selection 
& \textbf{Value} 
& $\sqrt{\epsilon_{\text{PEHE}}^{\text{in}}}$ & $\epsilon_{\text{ATE}}^{\text{in}}$ 
& $\sqrt{\epsilon_{\text{PEHE}}^{\text{out}}}$ & $\epsilon_{\text{ATE}}^{\text{out}}$ 
& Coverage & Selection \\
\midrule
0.1   & 0.678 & 0.096 & 0.709 & 0.112 & 0.93 & 0.74 
      & 0.1   & 0.286 & 0.009 & 0.288 & 0.010 & 0.96 & 0.94 \\
0.25  & 0.693 & 0.096 & 0.724 & 0.113 & 0.93 & 0.75 
      & 0.25  & 0.285 & 0.010 & 0.287 & 0.010 & 0.94 & 0.94 \\
\textbf{0.5} & \textbf{0.638} & \textbf{0.090} & \textbf{0.670} & \textbf{0.105} & \textbf{0.96} & \textbf{0.80}
      & \textbf{0.5} & \textbf{0.284} & \textbf{0.009} & \textbf{0.286} & \textbf{0.009} & \textbf{0.94} & \textbf{0.94} \\
1     & 0.712 & 0.103 & 0.746 & 0.115 & 0.96 & 0.79 
      & 1     & 0.285 & 0.013 & 0.287 & 0.014 & 0.94 & 0.92 \\
2.5   & 1.011 & 0.245 & 1.036 & 0.262 & 0.94 & 0.77 
      & 2.5   & 0.283 & 0.015 & 0.284 & 0.016 & 0.92 & 0.88 \\
\bottomrule
\end{tabular}
}
\end{table}

\begin{table}[t]
\centering
\caption{Sensitivity analysis of $\rho$ on IHDP and Twins datasets.}
\label{tab:sensitivity_rho}
\resizebox{\textwidth}{!}{
\begin{tabular}{c|cccccc|c|cccccc}
\toprule
\multicolumn{7}{c|}{\textbf{IHDP}} & \multicolumn{7}{c}{\textbf{Twins}} \\
\cmidrule(lr){1-7} \cmidrule(lr){8-14}
\textbf{Value} 
& $\sqrt{\epsilon_{\text{PEHE}}^{\text{in}}}$ & $\epsilon_{\text{ATE}}^{\text{in}}$ 
& $\sqrt{\epsilon_{\text{PEHE}}^{\text{out}}}$ & $\epsilon_{\text{ATE}}^{\text{out}}$ 
& Coverage & Selection 
& \textbf{Value} 
& $\sqrt{\epsilon_{\text{PEHE}}^{\text{in}}}$ & $\epsilon_{\text{ATE}}^{\text{in}}$ 
& $\sqrt{\epsilon_{\text{PEHE}}^{\text{out}}}$ & $\epsilon_{\text{ATE}}^{\text{out}}$ 
& Coverage & Selection \\
\midrule
10    & 0.698 & 0.108 & 0.735 & 0.123 & 0.96 & 0.78 
      & 10    & 0.299 & 0.015 & 0.306 & 0.015 & 0.92 & 0.62 \\
50    & 0.711 & 0.098 & 0.745 & 0.116 & 0.95 & 0.79 
      & 50    & 0.289 & 0.011 & 0.291 & 0.012 & 0.90 & 0.88 \\
\textbf{100} & \textbf{0.638} & \textbf{0.090} & \textbf{0.670} & \textbf{0.105} & \textbf{0.96} & \textbf{0.80} 
      & \textbf{100} & \textbf{0.284} & \textbf{0.009} & \textbf{0.286} & \textbf{0.009} & \textbf{0.94} & \textbf{0.94} \\
200   & 0.737 & 0.103 & 0.772 & 0.123 & 0.94 & 0.76 
      & 200   & 0.286 & 0.012 & 0.288 & 0.013 & 0.94 & 0.92 \\
1000  & 0.751 & 0.111 & 0.785 & 0.130 & 0.93 & 0.76 
      & 1000  & 0.284 & 0.010 & 0.285 & 0.011 & 0.92 & 0.94 \\
\bottomrule
\end{tabular}
}
\end{table}
\subsection{Model Implementation}
\label{imp_app}
We implement all models using PyTorch and optimize them with the Adam optimizer. The key hyperparameters include the size of each hidden layer, learning rate, the loss coefficients $\lambda_1$, $\lambda_2$, the penalty coefficient $\rho$, and the number of training epochs. These hyperparameters are manually tuned through empirical trials. The search ranges are as follows: hidden layer size in \(\{30, 40, 50, 60, 70\}\), learning rate in \(\{5 \times 10^{-4}, 10^{-3}, 2 \times 10^{-3}, 3 \times 10^{-3}\}\), \(\lambda_1, \lambda_2\) in \(\{0.1, 0.25, 0.5, 1, 2\}\), \(\rho\) in \(\{10, 50, 100, 200\}\), \text{and number of training epochs in} \(\{700, 800, 900, 1000, 1100\}\). 

\end{document}